\documentclass[10pt]{article} 
\usepackage[preprint]{tmlr}

\usepackage{amsmath,amsfonts,bm}

\def\eqref#1{equation~\ref{#1}}

\def\1{\bm{1}}

\DeclareMathAlphabet{\mathsfit}{\encodingdefault}{\sfdefault}{m}{sl}
\SetMathAlphabet{\mathsfit}{bold}{\encodingdefault}{\sfdefault}{bx}{n}

\usepackage{hyperref}
\usepackage{url}
\usepackage{cleveref}
\usepackage{adjustbox}
\usepackage{graphicx}
\usepackage{booktabs} 
\usepackage{lipsum}
\usepackage{arydshln}
\usepackage{paralist} 
\usepackage{subcaption} 
\usepackage{graphicx} 
\usepackage[most]{tcolorbox}
\usepackage{listings}
\usepackage{xcolor}
\usepackage{amsfonts}
\definecolor{headergray}{RGB}{245, 245, 245}
\definecolor{codebg}{RGB}{250, 250, 250}
\definecolor{interpreterbg}{RGB}{240, 240, 255} 
\title{Exploring Extrinsic and Intrinsic Properties for Effective Reasoning with Code Interpreter}

\author{\name Patomporn Payoungkhamdee \email patomporn.p\_s21@vistec.ac.th \\
      \addr Vidyasirimedhi Institute of Science and Technology
      \AND
      \name Napat Laosaengpha \email napat.r@vistec.ac.th \\
      \addr Vidyasirimedhi Institute of Science and Technology
      \AND
      \name Jenta Wonglertsakul \email jenta.w@live.ku.th \\
      \addr Kasetsart University
      \AND
      \name Pittawat Taveekitworachai \email pittawat@scb10x.com \\
      \addr SCB 10X
      \AND
      \name Pume Tuchinda \email pumet\_pro@vistec.ac.th \\
      \addr Vidyasirimedhi Institute of Science and Technology
      \AND
      \name Panjapong Poobanchuen \email panjapong.pooba@kmutt.ac.th \\
      \addr King Mongkut's University of Technology Thonburi
      \AND
      \name Ekapol Chuangsuwanich \email ekapol.c@chula.ac.th \\
      \addr Department of Computer Engineering \\ Chulalongkorn Univesity
      \AND
      \name Can Udomcharoenchaikit \email canu\_pro@vistec.ac.th \\
      \addr Vidyasirimedhi Institute of Science and Technology    
      \AND
      \name Samuel Cahyawijaya \email samuelcahyawijaya@cohere.com \\
      \addr Cohere
      \AND
      \name Peerat Limkonchotiwat \email peerat@aisingapore.org \\
      \addr AI Singapore
      \AND
      \name Sarana Nutanong \email snutanon@vistec.ac.th \\
      \addr Vidyasirimedhi Institute of Science and Technology      
}

\def\month{MM}  
\def\year{YYYY} 
\def\openreview{\url{https://openreview.net/forum?id=XXXX}} 

\begin{document}

\maketitle

\begin{abstract}
Reasoning with a Code Interpreter (CI) has emerged as an effective paradigm for enhancing the reasoning capabilities of large language models (LLMs) through executable computation and iterative verification. Despite its growing adoption, the behavioral properties underlying effective code reasoning remain largely underexplored. In this work, we investigate code reasoning from two distinct perspectives inspired by prior studies of natural language reasoning: extrinsic properties, represented by \emph{crucial tokens}, and intrinsic properties, represented by \emph{code-specific cognitive behaviors}. Across multiple LLMs, we find that stronger CI reasoning models consistently exhibit a higher prevalence of crucial tokens and cognitive behaviors, particularly verification, backtracking, and backward chaining. Building on these observations, we examine how these properties can be leveraged during both inference and training. At inference time, appending code-specific \emph{crucial tokens} improves performance on several reasoning capabilities, including mathematical, ordering, and optimization, while yielding limited benefits elsewhere. At training time, augmenting a state-of-the-art framework with code-specific cognitive behaviors improves supervised fine-tuning and reinforcement learning performance in two of three evaluated models. Further analysis shows that these behaviors reduce overthinking in incorrect responses and improve token efficiency, while also revealing factors that limit gains in a certain model. Our findings provide the first systematic characterization of effective reasoning with CI and demonstrate both the potential and limitations of leveraging key properties to improve CI-based reasoning.


%
\end{abstract}

\section{Introduction}

Reasoning with a Code Interpreter (CI)~\citep{pot,pal} has emerged as a promising paradigm to enhance the problem-solving capabilities of Large Language Models (LLMs). By delegating computation to an external execution environment while focusing on high-level reasoning~\citep{gou2024tora,r1-code}, CI-based reasoning mitigates common failures of natural language (NL) reasoning~\citep{wei2022cot,wang2023selfconsistency}, including arithmetic errors and brittle multi-step recursion. Prior work showed that reasoning with CI improves downstream performance, with gains of up to 13.5\% on grade-school mathematics benchmarks~\citep{cobbe2021gsm8k}. These advances underscore the importance of external tools for improving LLM reasoning and planning~\citep{react,codeact}. They are further reflected in coding agents, where models must execute, test, and iteratively debug code within feedback loops~\citep{swe-agent,AutoCodeRover}. In this setting, CI-based reasoning grounds model outputs from external evidence, reducing hallucinations and improving downstream performance \citep{RepairAgent}.

Concurrently, a growing body of research has sought to understand the mechanisms underlying NL reasoning. We organize this paper into two distinct perspectives: extrinsic and intrinsic. Extrinsic approaches examine how inference-time interventions, such as \emph{crucial token} \citep{epistemic-tokens,critical-tokens} injection, influence reasoning behavior. For example, \citet{s1} showed that appending a \textit{wait} token during inference encourages additional reasoning computation, improving performance in certain settings \citep{wu2025itssimpleanalysissimple,zhao2026testtimescalingreasoningmodels}.
For the intrinsic perspective, we define intrinsic properties as cognitive behaviors (\textsc{CoBe}) \citep{cognitive-behavior,satori,open-vision-reasoner}, such as planning, verification, and backward chaining, that emerge during training and correlate with higher task performance.

\begin{figure}[h!]
\centering\includegraphics[width=\textwidth]{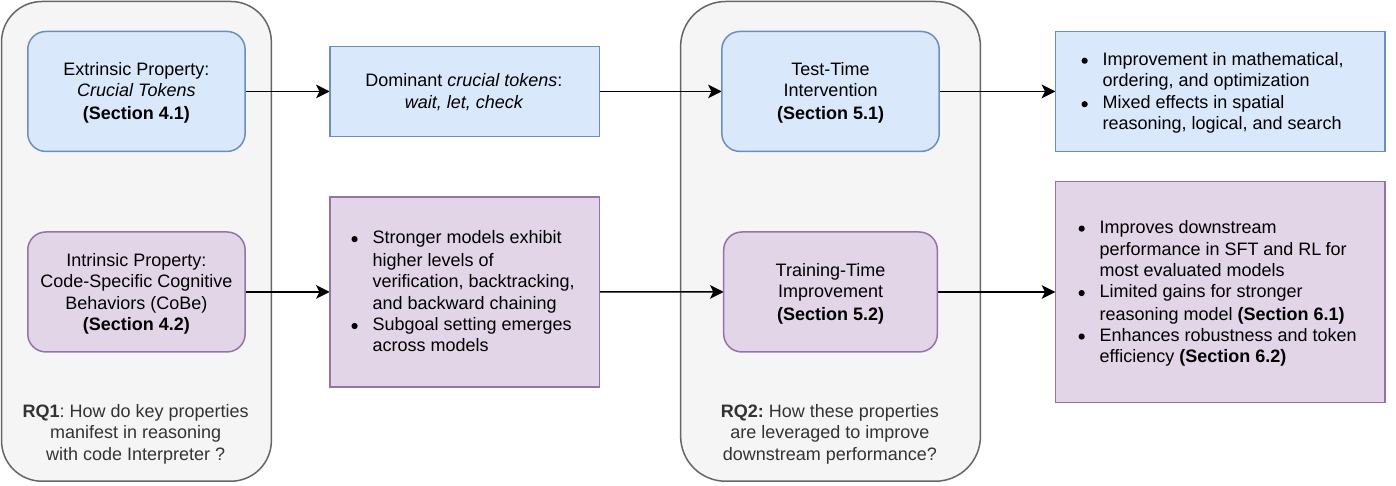} 
    \caption{
    Overview of how extrinsic and intrinsic properties characterize and improve reasoning with CI.
    }
    \label{fig:overall}
\end{figure}

%
Despite these extensive studies on NL reasoning, the key properties underlying effective CI-based reasoning remain underexplored. 
Our central question is: \textit{How do crucial tokens and \textsc{CoBe} manifest and leveraged in CI-based reasoning?}
As shown in \Cref{fig:overall}, this central question is decomposed into two research questions: 
\begin{compactitem}
\item \textbf{RQ1:} How do \emph{crucial tokens} (extrinsic) and \textsc{CoBe} (intrinsic cognitive behaviors) manifest in effective reasoning with the CI?
\item \textbf{RQ2:} How can \emph{crucial tokens} and \textsc{CoBe} be leveraged, at test time and training time, to improve downstream performance?
\end{compactitem}

To answer RQ1, we investigate CI-based reasoning across both properties. First, we examine \textbf{extrinsic properties} by identifying frequently occurring tokens across multiple LLMs, termed \emph{crucial tokens} (\Cref{subsec:extrinsic}). We find that the strongest model consistently exhibits higher frequencies of these tokens, notably \textit{wait}, \textit{let}, and \textit{check}.
Second, we analyze \textbf{intrinsic properties} by extending the cognitive behavior from \citep{cognitive-behavior} to CI-based reasoning (\Cref{subsec:intrinsic}). We observe that a high-performing model exhibits more code-specific behaviors, particularly checking intermediate results, revising unsuccessful attempts, and working backward from the target outcome.
In contrast, the behavior of decomposing a problem into smaller, manageable subgoals was consistently observed across all models, regardless of their performance level.

Building on these findings, we next investigate RQ2: how the identified properties can be leveraged to improve downstream performance at inference and training time.
At inference, we evaluate whether appending \emph{crucial tokens} improves performance on reasoning and planning tasks (\Cref{sec:test-time-scale-experiment}). We observe gains in mathematical, ordering, and optimization tasks, but mixed results on spatial, logical, and search reasoning, consistent with prior work \citep{wu2025itssimpleanalysissimple,zhao2026testtimescalingreasoningmodels} showing that test-time scaling is task-dependent.
At training time, we examine whether inducing code-specific cognitive behaviors improves CI-based reasoning (\Cref{sec:exploit-beh-experiment}). Incorporating \textsc{CoBe} into the recent CI-based training framework of \citet{retool} improves accuracy in both supervised fine-tuning and reinforcement learning stages in two of three models across both training stages. In particular, \textsc{CoBe} yields gains of 6.3\% during supervised fine-tuning and 10.7\% during reinforcement learning on Qwen2.5-7B-Instruct.
In contrast, for Qwen3-8B, the reduction of \emph{crucial token} frequency leads to lower performance relative to standard training, indicating that extrinsic properties also contribute to effective CI-based reasoning (\Cref{sec:analysis:qwen3-degrade}).
Further analyses in \Cref{sec:analysis-beyond-accuracy} indicate that \textsc{CoBe} mitigates overthinking and enhances token efficiency throughout the reinforcement learning process.

%
%

Our main contributions are summarized as follows:
\begin{compactitem}
\item We conduct a systematic study of effective reasoning with CI, identifying \emph{crucial tokens} and \emph{cognitive behaviors} and characterizing how these properties manifest in CI-based reasoning.
\item We show that these properties can be leveraged at both test time and training time, improving downstream performance through test-time scaling and training-time augmentation.
\item We analyze the benefits and limitations of these properties across models, revealing factors that improve efficiency and factors that limit their performance gains in a certain model.
\end{compactitem}

\section{Related Works}
\subsection{Reasoning with Code-Interpreter}

\citet{pot} and \citet{pal} introduced tool use through Code Interpreters, in which reasoning is conducted via program generation. By producing executable code and offloading computation to external environments, this paradigm allows LLMs to solve problems that require complex reasoning and precise calculations.
By grounding intermediate reasoning in execution, these approaches improve reliability on mathematical and algorithmic tasks \citep{wang2024mathcoder,gou2024tora,li2025torlscalingtoolintegratedrl}.
Subsequent work extended this paradigm to broader tool use. \citet{react} interleaved reasoning and tool invocation for grounded problem solving, while \citet{toolformer,jin2025searchr,li-etal-2025-search,wei-etal-2025-webagent,zheng-etal-2025-deepresearcher} enabled models to determine by themselves when and how to use external tools. Related studies have further explored tool-augmented agents operating in external environments, including coding agents that iteratively generate, execute, test, and refine code \citep{codeact,swe-agent,AutoCodeRover,RepairAgent,s-star-tts-code}.
More recently, reinforcement learning with verifiable rewards (RLVR) \citep{deepseekr1,olmo2026olmo3} has been applied to Code Interpreter reasoning. Frameworks such as \citet{retool} and \citet{r1-code} showed that outcome-based rewards improve tool interaction and mathematical reasoning through iterative execution and verification. While prior work primarily focuses on improving downstream performance, our work investigates the properties underlying effective Code Interpreter reasoning and how they can be leveraged at both inference and training time.

%
%
%
%

\subsection{Key Properties in Natural Language Reasoning}

To systematically investigate the properties of reasoning in CI-based systems, building on prior studies of natural language reasoning, we define the extrinsic property as interactions with \textit{crucial tokens} \citep{epistemic-tokens} and the intrinsic property as the dynamics of \textit{cognitive behaviors} \citep{cognitive-behavior}.

\paragraph{Extrinsic Properties.}
Early work on exploiting the extrinsic properties of extended reasoning in LLMs explored inference-time interventions that encourage models to continue reasoning after reaching a completion signal, such as repeatedly appending a \textit{wait} token \citep{s1}. Although this simple strategy shows performance improvements on mathematical benchmarks, subsequent studies questioned its robustness and generalization. \cite{wu2025itssimpleanalysissimple} showed that gains from repeated \textit{wait} injections are highly inconsistent, with models often oscillating between answers or producing nearly identical reasoning traces despite additional computation. Extending the analysis to knowledge-intensive tasks, \cite{zhao2026testtimescalingreasoningmodels} found no consistent benefit from budget forcing and reported increased hallucination rates for stronger reasoning models, while \cite{ghosal2025doesthinkinghelpmirage} observed a non-monotonic scaling effect in which excessive deliberation can degrade performance through overthinking. More recent work argues that test-time scaling gains depend less on reasoning length and more on the informational content of reasoning trajectories. The 80:20 hypothesis suggests that only a small fraction of reasoning tokens contribute disproportionately to performance improvements \citep{80-20}, while \cite{critical-tokens} identifies ``critical tokens'' that coincide with key reasoning transitions and carry outsized influence on final predictions. Similarly, \cite{epistemic-tokens} argues that improvements stem from epistemic verbalization, the explicit expression of uncertainty, doubt, and self-correction, rather than from specific trigger tokens or longer chains of thought alone. 
However, these studies focus predominantly on natural language reasoning, leaving the extrinsic properties of reasoning with external tools, particularly code interpreters, largely unexplored.

\paragraph{Intrinsic Properties.}
A growing line of research examines the intrinsic cognitive behaviors and emergent exploration tendencies of LLMs that dictate their capacity for reinforcement learning (RL) and test-time scaling. An influential study in this area, \cite{cognitive-behavior} empirically demonstrated that foundational cognitive habits, such as verification, subgoal setting, backtracking, and backward chaining, serve as important priors for RL-driven self-improvement, while models exhibiting these behaviors scale effectively under RL, behavior-deficient models quickly plateau unless primed with behavior-rich trajectories. Building on this premise, recent works focus on how to systematically instill, condense, or manipulate these internal mental mechanics. \cite{SkillFactory} introduces a self-distillation framework to isolate and transmit these pure structural reasoning behaviors to smaller base models to prevent early RL stagnation, while \cite{sot} draws from human cognitive psychology to compress verbose reasoning traces into efficient, dense internal intermediaries via adaptive sketching. To maximize the utility of these habits during inference, \cite{CREST} identifies specialized internal attention heads dedicated to non-linear reasoning steps and introduces test-time steering mechanisms to balance overthinking and underthinking. Furthermore, models like Satori \citep{satori} manifest these intrinsic behaviors explicitly by framing verification and backtracking as autoregressive ``meta-actions'' within a unified chain-of-action-thought. Extending this paradigm beyond pure text, \citet{vigorl,open-vision-reasoner} demonstrate that these linguistic cognitive habits structurally generalize to visual reasoning through iterative grounding and verification under RLVR. Collectively, these studies showed that a model's intrinsic cognitive architecture governs its learning efficiency. However, these dynamics remain underexplored in tool-integrated environments like reasoning with CI, where external compiler feedback introduces distinct, deterministic feedback loops.

\section{Background}

In this section, we discuss reasoning in LLMs, covering NL reasoning and reasoning with CI interactions, including multi-turn settings (\Cref{sec:reason-ci}). We then examine how test-time scaling can be integrated into the CI-based reasoning framework (\Cref{sec:test-time-scale}). Finally, \Cref{sec:sft-and-rl} presents the training pipeline, including supervised fine-tuning for cold-start initialization and reinforcement learning for LLMs' reasoning with CI.

\subsection{Reasoning with Code Interpreter}
\label{sec:reason-ci}

Natural language reasoning considers an input problem $x$ and a generated response $\hat{y}$, where the response is sampled from a model policy $\pi_\theta$ parameterized by $\theta$, i.e., $\hat{y} \sim \pi_\theta(\cdot \mid x)$.
The generated response typically consists of a reasoning trajectory (rationale) $\hat{r}$ and a final prediction $\hat{a}$, where the answer is commonly formatted within a predefined pattern, such as \texttt{\textbackslash boxed}\{\textit{answer}\}.

Unlike traditional natural language reasoning, where a model has to tackle both high-level problem-solving and actual computation, reasoning with CI lets the model generate code, commonly Python, to leverage the computational capability from CI, letting the model focus solely on problem-solving.
Conventionally, for the first turn, the model has to generate both reasoning trajectory $\hat{y}$ and code $\hat{c}$ from a given problem, represented as $(\hat{y}_0,\hat{c}_0) \sim \pi_\theta(\cdot|x)$.
After that, the code will be parsed by CI, resulting in an output $\hat{o}_0\sim \text{CI}(\hat{c}_0)$. The model then takes code execution output and generate the next turn $(\hat{y}_1,\hat{c}_1) \sim \pi_\theta(\cdot|\hat{o}_0,\hat{y}_0,\hat{c}_0, x)$, typically called multi-turn interaction with CI until reaching the final turn ($T$), generally represented as 
\begin{equation}
    (\hat{y}_T,\hat{c}_T) \sim \pi_\theta(\cdot|\mathbf{\hat{o}}_{t<T}\mathbf{\hat{y}}_{t<T},\mathbf{\hat{c}}_{t<T}, x)
    \label{eq:multiturn-gen}
\end{equation}
Similar to typical reasoning where $\hat{y}_T$ contains rationale $\hat{r}$ and final answer $\hat{a}$.
A representative trace of this structure is illustrated in \Cref{fig:reason-with-ci}.


\subsection{Test-Time Scaling}
\label{sec:test-time-scale}

Test-time scaling is an emerging paradigm that improves downstream performance by allocating additional computational resources during inference \citep{s1,wang2023selfconsistency}. Existing approaches can be broadly categorized into parallel and sequential scaling \citep{zhang2025surveytesttimescalinglarge}. Parallel scaling generates multiple independent reasoning trajectories and aggregates or selects among them, whereas sequential scaling devotes computation to extending a single trajectory, enabling iterative refinement and error correction.
To study how models interact with extrinsic factors, we focus on sequential test-time scaling in the context of CI-based reasoning and investigate its scaling behavior across different capabilities and model variants.

To investigate these dynamics, we manipulate the test-time scaling budget by overriding the model’s self-termination signal. Typically, reasoning models signify the completion of a reasoning by generating a specific token, such as \textsc{</think>}. We intervene by replacing this token with a \emph{crucial token} ($w$), such as \textit{wait}, which compels the model to extend its reasoning. Formally, given an initial reasoning $\hat{r}_0 \sim \pi_\theta(\cdot|x)$, we iteratively inject $N$ such tokens to produce an extended trajectory:
\begin{equation}
\hat{r}_N \sim \pi_\theta(\cdot \mid w, \hat{r}_{N-1}, \dots, w, \hat{r}_0, x)
\end{equation}
The final code, $\hat{c}$, is then sampled conditioned on the full extended reasoning:
\begin{equation}
\hat{c} \sim \pi_\theta(\cdot \mid \hat{r}_N, w, \hat{r}_{N-1}, \dots, w, \hat{r}_0, x)
\end{equation}
This methodology allows us to systematically analyze the scaling behavior of model accuracy as a function of the total tokens allocated to the reasoning phase, forcing the model to further explore the problem space or verify its logic before committing to a computational solution.

\subsection{Supervised Fine-Tuning and Reinforcement Learning}
\label{sec:sft-and-rl}

Following prior work that studied CI-based reasoning with LLMs \citep{retool,r1-code}, the training process is divided into two main phases. The first phase uses supervised fine-tuning (SFT) as a cold start to help the model understand the basic structure of multi-turn CI tasks.
In the second phase, reinforcement learning (RL) is used to further optimize the model through multi-turn interaction with the CI. During this stage, the model learns by writing code and interacting with a real CI environment over several turns. This enables the model to delegate compute-intensive operations, such as arithmetic and recursive computations, to the CI while focusing on higher-level reasoning before producing a final solution.

\paragraph{Supervised Fine-Tuning.}
In cold-start SFT for multi-turn CI interactions, the model is typically trained on a curated dataset denoted as $\mathcal{D}^\text{SFT} = \{(x_i, \mathbf{y}_i, \mathbf{c}_i, \mathbf{o}_i)\}_{i=1}^N$, where $N$ represents the total number of training samples. Each individual sample $X_i$ is a tuple consisting of a natural language instruction $x_i$, a gold-standard response $\mathbf{y}_i$, the corresponding code snippets $\mathbf{c}_i$, and the resulting execution outputs $\mathbf{o}_i$.
To ensure the model focuses its learning capacity on generative tasks, specifically, the text responses and executable code. The training objective utilizes a masked loss function.
For a given training sample $i$, the SFT loss over a sequence of $J$ tokens is formulated as:
\begin{equation}
    \mathcal{L}_i^\text{SFT} = - \frac{1}{\sum_{j} \delta_{i,j}} \sum_{j=1}^J \delta_{i,j} \log P_\theta(X_{i,j} | X_{i,<j})
\end{equation}
In this objective, a delta function $\delta_{i,j}$ serves as a binary mask, where $\delta_{i,j} = 1$ if the token $X_{i,j}$ belongs to the set of target outputs $\{\mathbf{y}_i, \mathbf{c}_i\}$, and $\delta_{i,j} = 0$ otherwise. This approach effectively isolates the gradient updates to the model's primary generative tasks.

\paragraph{Reinforcement Learning.}
Following the ReTool framework \citep{retool}, the RL pipeline employs the Proximal Policy Optimization (PPO) paradigm \citep{ppo}. In this specific implementation, Group Relative Policy Optimization (GRPO) \citep{grpo} is utilized to refine the policy. 
The RL training dataset, denoted as $\mathcal{D}^\text{RL}$, comprises pairs of problem ($x$) and their corresponding ground-truth answers ($a$). The optimization objective for the policy update is defined as:
\begin{equation}
\resizebox{\columnwidth}{!}{$
    \mathcal{J}^{\mathrm{GRPO}}(\theta)=
    \frac{1}{G}\sum_{g=1}^G\frac{1}{|\hat{y}_g|}\sum_{j=1}^{|\hat{y}_g|}
    \left\{\min\left[\frac{\pi_{\theta}(\hat{y}_{g,j}\mid x,\hat{y}_{g,<j};\text{CI})}{\pi_{\theta_{\mathrm{old}}}(\hat{y}_{g,j}\mid x,\hat{y}_{g,<j};\text{CI})}\hat{A}_{g,j},\operatorname{clip}\left(\frac{\pi_{\theta}(\hat{y}_{g,j}\mid x,\hat{y}_{g,<j};\text{CI})}{\pi_{\theta_{\mathrm{old}}}(\hat{y}_{g,j}\mid x,\hat{y}_{g,<j};\text{CI})},1-\varepsilon,1+\varepsilon\right)\hat{A}_{g,j}\right]\right\}
$}
\end{equation}

In this formulation, the objective maximizes the lower bound of the clipped surrogate advantage. The probability ratio between the current policy $\pi_\theta$ and the behavior policy $\pi_{\theta_{\text{old}}}$ is constrained by the hyperparameter $\varepsilon$ to ensure stable updates, while the advantage estimator $\hat{A}_{g,j}$ guides the policy toward sequences that yield higher relative rewards within the CI environment.

In alignment with the ReTool framework, the reward signal is formulated as a binary outcome contingent on the accuracy of the final derived answer. This sparse reward mechanism incentivizes the model to converge on the correct solution via valid reasoning and code execution. Formally, the reward function $R(a, \hat{a})$ is defined such that $R(a, \hat{a}) = 1$ if the predicted final answer $\hat{a}$ matches the ground-truth label $a$, and $-1$ otherwise.
Within the GRPO framework, the advantage $\hat{A}_{g,j}$ for a given output sequence is determined by evaluating its reward relative to the mean performance of a group of $G$ sampled outputs for the same query. Specifically, for each trajectory $o_g$ within a sampled group $\{o_1, o_2, \dots, o_G\}$, the advantage is computed as the z-score of the group's reward distribution $R(x,\hat{y}_g)$. This relative advantage estimation utilizes the group's collective performance as a dynamic baseline, eliminating the need for a separate value function (critic) and thereby streamlining the RL process for CI-based tasks.


\section{Key Properties of Effective Reasoning with Code Interpreter (RQ1)}
\label{sec:key-properties}

In this section, we decompose RQ1 into two orthogonal lines of investigation. First, we investigate the extrinsic properties of effective reasoning with CI by analyzing the 
frequency of specific \textit{crucial tokens} generated by strong models compared to weaker baselines in code-generation settings (\Cref{subsec:extrinsic}).
Second, we explore the intrinsic properties by quantifying cognitive behaviors specific to reasoning with CI (\Cref{subsec:intrinsic}), offering empirical insights into the models' reasoning behavioral patterns.

\subsection{Extrinsic Property}
\label{subsec:extrinsic}


\paragraph{Motivation.}
Test-time scaling via token injection has been shown to improve NL reasoning in certain settings \citep{s1}, although its effectiveness varies across models \citep{wu2025itssimpleanalysissimple,zhao2026testtimescalingreasoningmodels}. At the same time, code interpreters have demonstrated strong benefits for complex reasoning tasks \citep{pot,retool}, yet their surface-level reasoning patterns remain largely unexplored. We therefore investigate \textit{crucial tokens} in CI-based reasoning and evaluate their utility for test-time scaling.



\paragraph{Setup.}
Similar to \citet{epistemic-tokens}, we evaluate three distinct model families: Qwen3-8B \citep{qwen3} as a representative reasoning model, Qwen2.5-7B-Instruct \citep{qwen25} as a general-purpose baseline, and include Llama3.1-8B-Instruct \citep{llama3} as a standard instruction-tuned model.
Experiments are conducted on SymBench \citep{symbench}, a comprehensive benchmark designed for symbolic reasoning. This dataset encompasses a diverse set of tasks, including mathematics, spatial reasoning, logic, sequential ordering, optimization, and search-based problems.
Building on the concept of epistemic tokens\footnote{\textit{wait, hmm, perhaps, maybe, alternatively, guess, correct, check, so, now, actually, likely, might,} and \textit{seems}.} \citep{epistemic-tokens} originally identified in NL reasoning, we empirically investigate their surface patterns for reasoning with CI. Furthermore, we include \textit{let}, an auxiliary definition in programming language \citep{next700programlang}.
Collectively, we refer to these as \textit{crucial tokens}.
To quantify their prevalence, we measure the frequency of each \emph{crucial token} normalized by the total number of generated tokens.

\begin{figure}[h!]
    \centering
    \includegraphics[width=\textwidth]{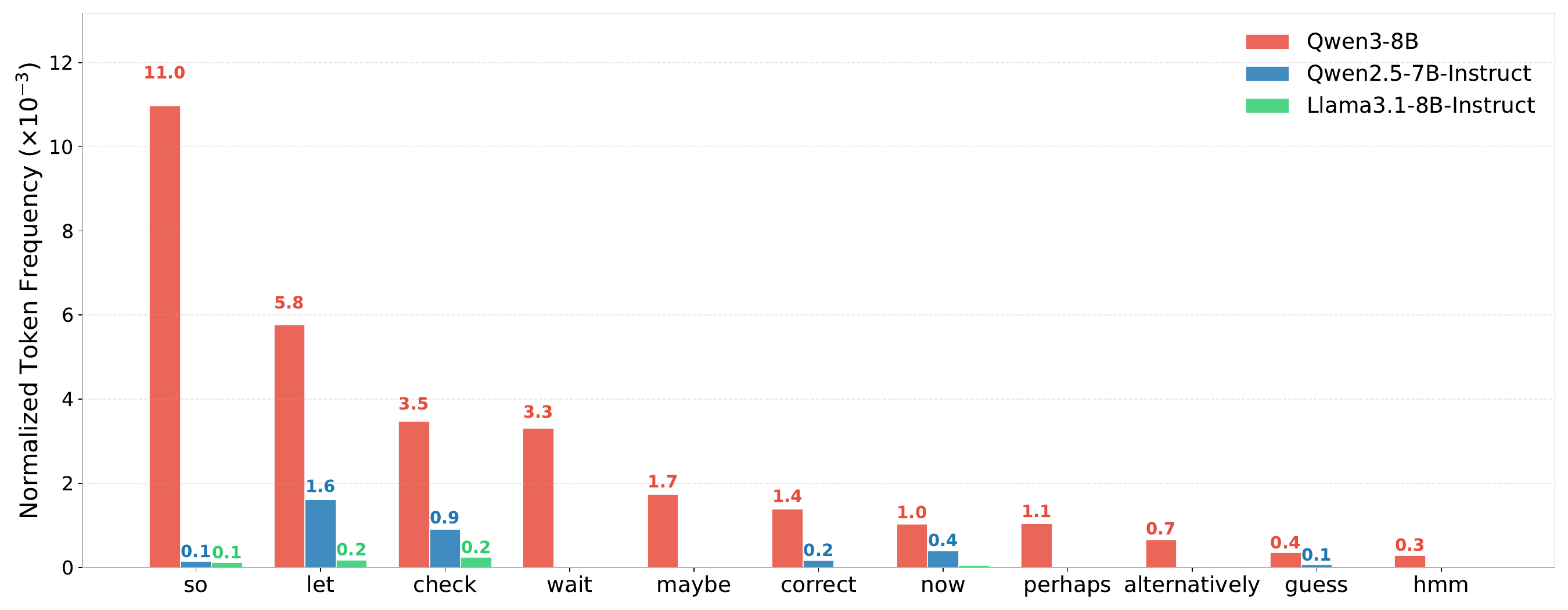} 
    \caption{
    Distribution of \emph{crucial token} frequencies across various models on SymBench. We present the ten most frequent epistemic tokens identified by \citet{epistemic-tokens}, along with the inclusion of the \textit{let} token.
    }
    \label{fig:ngram-comparison}
\end{figure}

\paragraph{Result and Discussion.}
Regarding model performance, a high-capacity model such as Qwen3 achieves an accuracy of 66.5\%, whereas Qwen2.5 and Llama3.1 attain 36.2\% and 10.3\%, respectively.
Focusing on Qwen3-8B, which is the best-performing model, \Cref{fig:ngram-comparison} shows that the four most frequent \emph{crucial tokens} are \textit{so, let, check, and wait}, in this order, which is consistent with prior studies \cite{80-20,critical-tokens,epistemic-tokens}. Notably, \textit{let} appears as the second most dominant token for reasoning with CI.
In contrast, the frequency of \emph{crucial tokens} is substantially lower in Qwen2.5 and Llama3.1. Nevertheless, \textit{let} and \textit{check} remain the two most frequent tokens, suggesting that they may play a significant role in inference-time reasoning with CI.
Building on these observations, \Cref{sec:test-time-scale-experiment} evaluates the utility of \emph{crucial tokens} for inference-time scaling. We focus on \textit{let} and \textit{check} due to their prevalence across top-performing models, and \textit{wait} for comparing with prior works, excluding \textit{so} because it is largely absent from the second-best model.

\subsection{Intrinsic Property}
\label{subsec:intrinsic}


\paragraph{Motivation.}
Prior work demonstrates that high performing models exhibit key cognitive behaviors, including verification, backtracking, subgoal setting, and backward chaining, which are largely absent in weaker models \citep{cognitive-behavior}. These behaviors enable models to validate intermediate results, revise unsuccessful reasoning paths, decompose complex problems into manageable subproblems, and reason backward from desired outcomes. Motivated by these findings, we investigate whether such behaviors also emerge in CI-based reasoning and the extent to which they contribute to solving challenging tasks.


\paragraph{Setup.}
Following \Cref{subsec:extrinsic}, we employ the same set of models and benchmark for our empirical analysis.
To investigate internal reasoning processes, we adopt an LLM-as-a-Judge framework, extending the approach of \citet{cognitive-behavior}.
However, preliminary experiments reveal that the original prompts, designed to evaluate cognitive behavior in natural language reasoning, do not consistently yield reliable outcomes in code-oriented reasoning contexts.
We therefore refine the probing prompt to better suit scenarios involving reasoning with a code interpreter (see \Cref{appendx:probing-code} for further details).
For the Oracle LLM, we use Qwen3-Next-80B-A3B-Instruct \citep{qwen3}, which has demonstrated strong performance on mathematical problem-solving and coding tasks.

\begin{figure}[h!]
    \centering
    \includegraphics[width=0.7\textwidth]{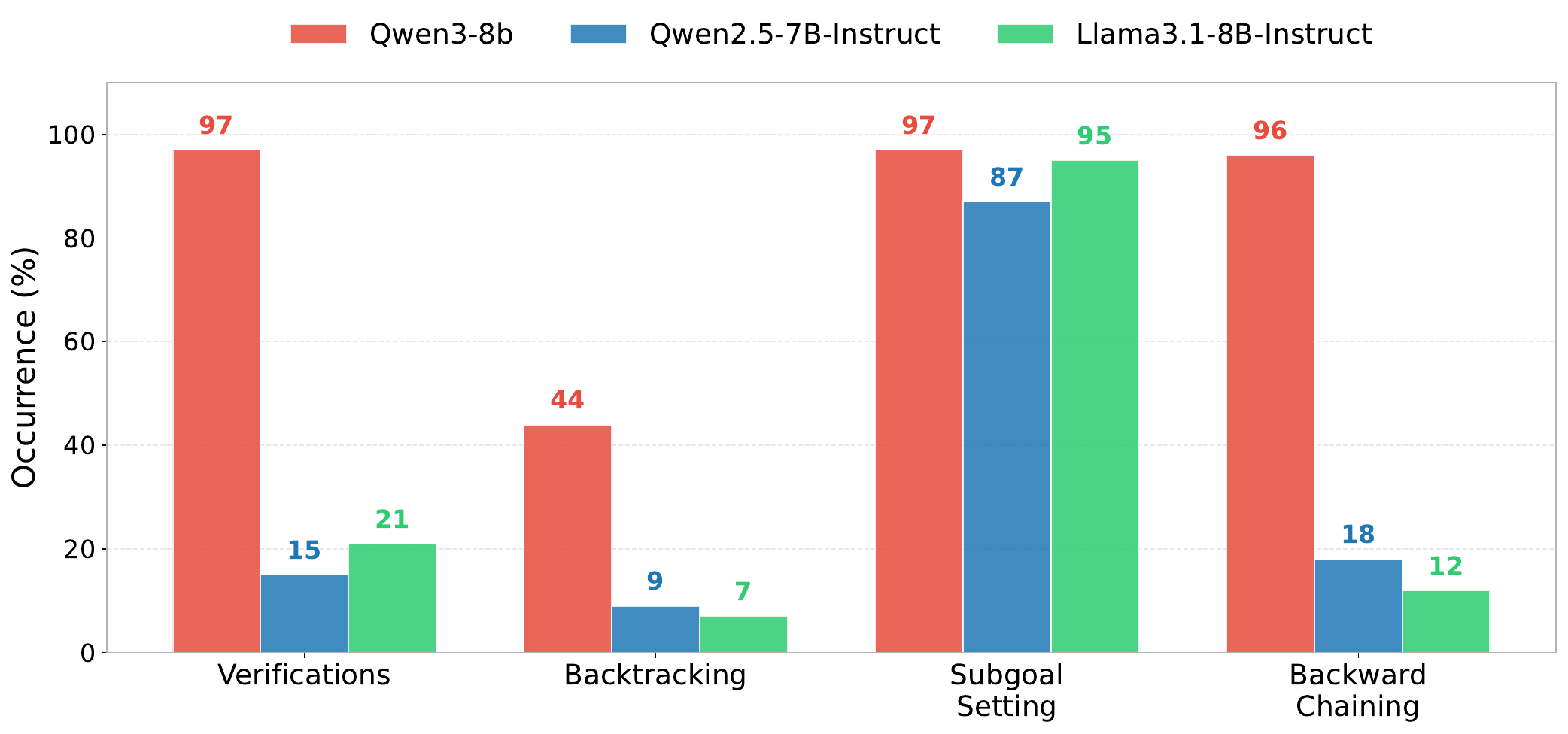}
    \caption{
        Analysis of reasoning behaviors across base models. The figure shows the percentage appearance of reasoning responses that exhibit key cognitive behaviors across comparable model sizes.
     }
    \label{fig:cognitive-beh}
\end{figure}

\paragraph{Result and Discussion.}
\Cref{fig:cognitive-beh} shows that Qwen3 exhibits the highest frequency across all cognitive behavior types. In particular, verification, subgoal setting, and backward chaining appear in at least 96\% of its responses. Backtracking is less frequent in Qwen3 (44\%), though still markedly higher than in Qwen2.5 and Llama3.1 (9\% and 7\%, respectively).
This pattern contrasts with typical natural language reasoning, where backtracking is most prevalent. It is the least frequent behavior in reasoning with CI. In contrast, subgoal setting dominates in our settings, where planning is implicitly encouraged, thereby increasing its prevalence.
The overall trend is consistent with \citet{cognitive-behavior}. These findings suggest that such cognitive behaviors are also foundational for solving challenging problems in reasoning with CI, as reflected in downstream performance. Building on this insight, we incorporate these behaviors into an existing state-of-the-art training framework, designed for reasoning with CI, as detailed in \Cref{sec:exploit-beh-experiment}.

\section{Leveraging Key Properties To Enhance Downstream Performance (RQ2)}

Building upon the findings in \Cref{sec:key-properties}, we empirically examine the potential benefits and limitations of leveraging these properties in downstream applications.
For test-time scaling, \Cref{sec:test-time-scale-experiment} investigates the impact of three selected \emph{crucial tokens} identified in \Cref{subsec:extrinsic}. We assess their scalability and effectiveness across six capabilities on challenging tasks using a diverse range of models.
For training-time enhancement, inspired by the observations in \Cref{subsec:intrinsic} that cognitive behaviors are important for code reasoning, \Cref{sec:exploit-beh-experiment} integrates four cognitive reasoning behaviors into the training process. We further show that integrating these behaviors leads to further gains during both supervised fine-tuning and reinforcement learning.

\subsection{Test-Time Scaling with Crucial Tokens.}
\label{sec:test-time-scale-experiment}

\paragraph{Motivation.}
Sequential test-time scaling for NL reasoning has been extensively studied across diverse settings \citep{s1,wu2025itssimpleanalysissimple,zhao2026testtimescalingreasoningmodels}. 
However, its effectiveness and limitations in CI-based reasoning remain under-investigated. 
To address this, we conduct an empirical study of scaling behavior under \emph{crucial token} intervention and analyze performance variations across models.

\paragraph{Setup.}
For test-time scaling in reasoning with CI, we follow the implementation introduced in \Cref{sec:test-time-scale}. Based on the analysis in \Cref{subsec:extrinsic}, we select the tokens \textit{wait}, \textit{let}, and \textit{check} as the \emph{crucial tokens} for study in this section.
We adopt SymBench as the evaluation benchmark to analyze downstream scaling behavior across six reasoning capabilities. 
Our primary model is Qwen3-8B, and we additionally include its smaller counterpart, Qwen3-4B, to investigate the effect of model scale. 
To further examine the variance across model families, we incorporate DeepSeek-R1 \citep{deepseekr1} distilled models of comparable size, namely DeepSeek-R1-Distill-Qwen-7B and DeepSeek-R1-Distill-Llama-8B.

\begin{figure}[h!]
    \centering
    \includegraphics[width=\linewidth]{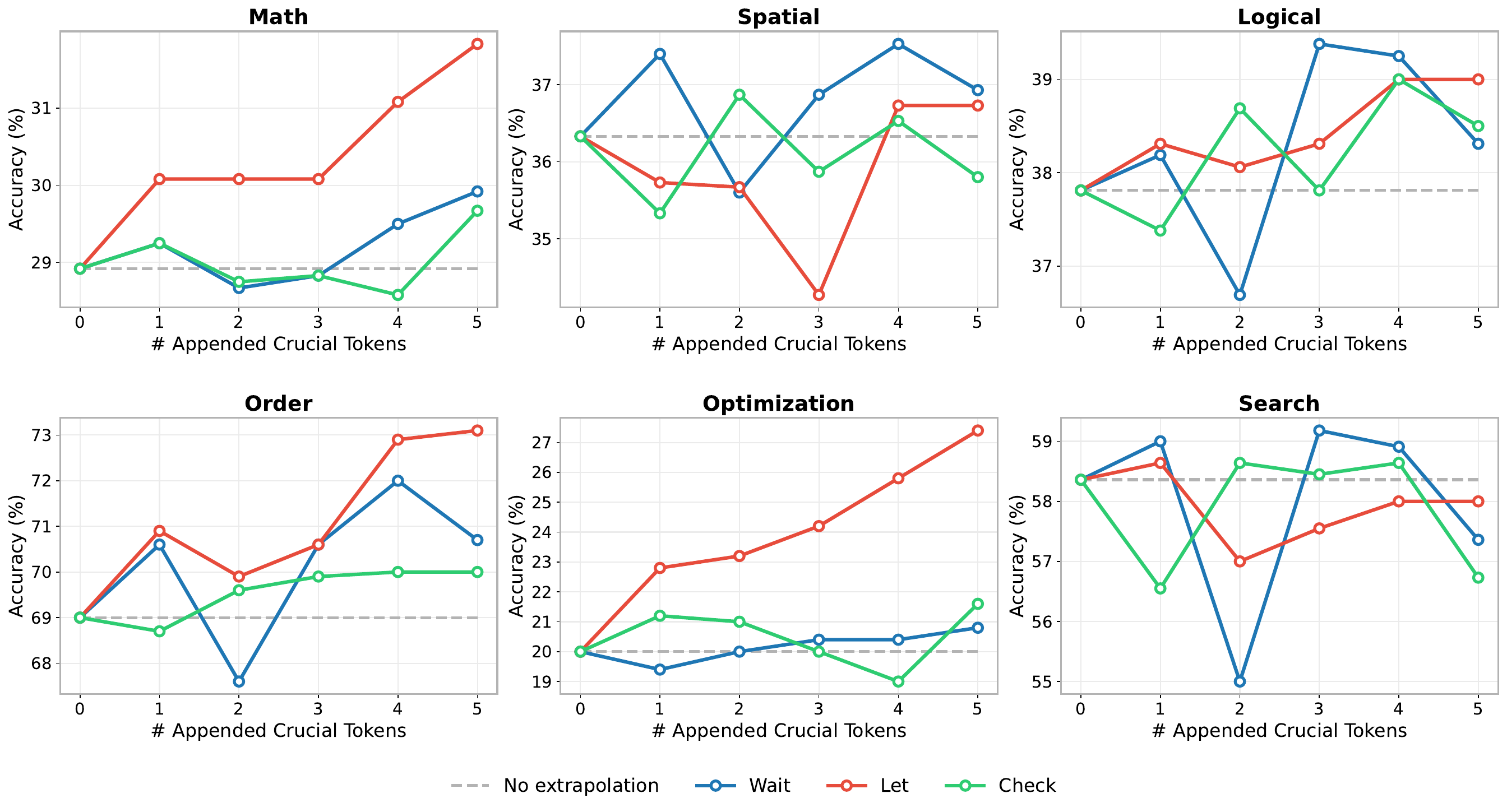}
    \caption{
        Accuracy progression of Qwen3-8B under test-time scaling, achieved by iteratively appending the \textit{crucial tokens}. Each marker denotes an iteration where the individual token is injected, corresponding to the generated token count shown on the x-axis.
     }
    \label{fig:scaling-trend}
\end{figure}

\paragraph{Result and Discussion.}
\Cref{fig:scaling-trend} shows that test-time scaling via the forced appending of \emph{crucial tokens} generally leads to substantial performance improvements over the non-extrapolated baseline across all evaluated reasoning domains.
The scaling effect is particularly pronounced and consistent in mathematical reasoning, order reasoning, and optimization tasks, where the \textit{let} token demonstrates near-monotonic gains and significantly outperforms alternative tokens as the number of appended tokens increases.
By contrast, although some tasks benefit from test-time scaling, their performance trajectories exhibit greater volatility and non-monotonic behavior, especially under the \textit{wait} and \textit{check} strategies. For example, in the search and logical capabilities, the \textit{wait} token produces sharp performance drops at intermediate token lengths before partially recovering, indicating sensitivity to the extrapolated sequence length.
Despite these task-dependent variations, the overall results suggest that the \textit{let} token provides the most reliable outcome for leveraging additional inference-time computation, yielding consistent improvements across most capabilities.
These findings are partially consistent with \citet{s1}, which reported that selecting an appropriate \emph{crucial token} under specific configurations can systematically enhance model performance, as observed for the \textit{let} token in mathematical reasoning.
Furthermore, our results indicate that, in certain capabilities such as spatial reasoning, logical reasoning, and search, simply appending \emph{crucial tokens} does not necessarily guarantee superior performance. This observation aligns with prior studies \citep{wu2025itssimpleanalysissimple,zhao2026testtimescalingreasoningmodels}, which similarly highlight the challenges when adopting test-time scaling in more complex reasoning settings.

Following the ablation setup in \citet{s1}, we inject individual \emph{crucial tokens} to evaluate their effects across different reasoning model families and token variants. 
As shown in \Cref{tab:scaling}, the effectiveness of \emph{crucial token} extrapolation varies substantially across model families and scales. 
For Qwen3-8B, the \textit{let} and \textit{check} tokens generally improve downstream performance across multiple reasoning domains, with \textit{let} achieving the best gains in mathematics and optimization, while \textit{check} performs best in spatial, logical, and search tasks. In contrast, the \textit{wait} token consistently underperforms the baseline. 
For the smaller Qwen3-4B model, \emph{crucial token} intervention yields limited gains, with degradation on most tasks and only marginal improvement in optimization, consistent with \citet{ghosal2025doesthinkinghelpmirage}, where sub-7B models struggle with test-time scaling on challenging problems.
The DeepSeek distilled models exhibit stronger scaling behavior. DeepSeek-R1-Distill-Llama-8B shows improvements primarily in mathematics, spatial, and logical reasoning, whereas DeepSeek-R1-Distill-Qwen-7B demonstrates consistent gains across all evaluated competencies, with substantial improvements in mathematics, spatial reasoning, and optimization under all token variants. 
Overall, these results suggest that the effectiveness of \emph{crucial token} intervention depends strongly on model architecture and scale. The most consistent gains are observed in mathematics and optimization, where Qwen3-8B and both DeepSeek distilled models benefit from at least one token variant, whereas improvements in spatial, logical, and search reasoning are more model-dependent.

\begin{table}[h!]
    \caption{Test-time scaling performance under different \emph{crucial token} injection strategies, evaluated reasoning models, Qwen3 in two sizes, and two Deepseek's distilled reasoning models.}
    \centering
    \begin{adjustbox}{max width=0.8\textwidth}
    \begin{tabular}{l|cccccc}
    \toprule
    \textbf{Method} & Math & Spatial & Logical & Order & Optimization & Search \\
    \midrule
    \multicolumn{1}{l|}{\underline{Qwen3-8B}} &
    \multicolumn{6}{c}{} \\
    No extrapolation & 28.92 & 36.33 & 37.81 & 69.00 & 20.00 & 58.36 \\
    2x \textit{wait} & 28.67 & 35.60 & 36.69 & 67.70 & 20.00 & 55.00 \\
    2x \textit{let} & \textbf{30.08} & 35.67 & 38.06 & \textbf{69.90} & \textbf{23.20} & 57.00 \\
    2x \textit{check} & 28.75 & \textbf{36.87} & \textbf{38.69} & \textbf{69.90} & 21.00 & \textbf{58.64}  \\
    \midrule
    \multicolumn{1}{l|}{\underline{Qwen3-4B}} &
    \multicolumn{6}{c}{} \\
    No extrapolation & \textbf{32.92} & \textbf{37.33} & \textbf{35.56} & \textbf{68.40} & 15.60 & \textbf{55.64} \\
    2x \textit{wait} & 28.92 & 35.67 & 33.19 & 65.30 & 10.60 & 53.36 \\
    2x \textit{let} & 30.00 & 34.33 & 33.75 & 64.00 & \textbf{16.60} & 52.00 \\
    2x \textit{check} & 29.17 & 33.80 & 33.88 & 64.90 & 16.20 & 51.64  \\
    \midrule
    \multicolumn{1}{l|}{\underline{DeepSeek-R1-Distill-Llama-8B} } &
    \multicolumn{6}{c}{} \\
    No extrapolation & 34.83 & 23.23 & 31.88 & \textbf{37.50} & 11.40 & \textbf{33.73} \\
    2x \textit{wait} & \textbf{35.25} & 22.47 & \textbf{34.94} & 35.50 & 11.60 & 33.36 \\
    2x \textit{let} & 33.92 & 21.33 & 34.19 & 36.30 & \textbf{13.00} & 31.36 \\
    2x \textit{check} & 35.17 & \textbf{24.73} & 32.19 & 37.00 & 11.40 & 32.91 \\
    \midrule
    \multicolumn{1}{l|}{\underline{DeepSeek-R1-Distill-Qwen-7B} } &
    \multicolumn{6}{c}{} \\
    No extrapolation & 18.58 & 19.07 & 23.81 & 36.00 & 3.60 & 27.64 \\
    2x \textit{wait} & 23.25 & 23.13 & 25.19 & \textbf{42.60} & 9.00 & \textbf{31.18} \\
    2x \textit{let} & \textbf{24.58} & \textbf{23.67} & \textbf{25.31} & 42.30 & 8.60 & 30.45 \\
    2x \textit{check} & 23.92 & 23.47 & 24.38 & 40.80 & \textbf{9.60} & 29.36 \\
    \bottomrule
    \end{tabular}
    \end{adjustbox}
    \label{tab:scaling}
\end{table}

\subsection{Incorporating Cognitive Behaviors at Training Time}
\label{sec:exploit-beh-experiment}

\textbf{Motivation.}
Inspired by \cite{cognitive-behavior}, prior work has shown that cognitive behaviors such as verification, backtracking, subgoal setting, and backward chaining can substantially improve the reasoning capabilities of weaker models.
In addition, our empirical findings in \Cref{subsec:intrinsic} suggest that these behaviors constitute important intrinsic properties for effective reasoning.
Motivated by these observations, we investigate whether equipping models with code-specific cognitive behaviors (\textsc{CoBe}) can further enhance downstream performance on reasoning tasks involving multi-turn interactions with code interpreter feedback.

\paragraph{Setup.}
To evaluate the effectiveness of \textsc{CoBe}, we adopt the training framework described in \Cref{sec:sft-and-rl}, which consists of SFT for cold-start initialization in a CI environment, followed by multi-turn reinforcement learning.
SFT To incorporate \textsc{CoBe} in the SFT pipeline, we develop a data synthesis pipeline extending the method of \citet{cognitive-behavior}. Given an existing reasoning trajectory $\mathbf{\hat{y}}$ and a cognitive behavior prompt $p_\text{CoBe}$, we generate an augmented trajectory $\mathbf{y}_\text{CoBe} \sim \text{LLM}(\mathbf{x}, \mathbf{y};p_\text{CoBe},\text{CI})$.
Following \citet{cognitive-behavior}, we employ Claude Sonnet 4.5 as the oracle language model for trajectory synthesis, motivated by its near-perfect performance on AIME 2025.\footnote{\url{https://www-cdn.anthropic.com/963373e433e489a87a10c823c52a0a013e9172dd.pdf}}
For consistency with ReTool, we primarily evaluate the Qwen2.5 family, including Qwen2.5-7B-Instruct-Coder and its instruction-tuned variants, and additionally consider Qwen3-8B, a reasoning model.
We evaluate on AIME24, AIME25, and MATH500 \citep{math500}.
Additional details are provided in \Cref{appendix:syn-cobe}.

\begin{table}[h!]
    \caption{Main results on mathematical problems.}
    \begin{center}
    \begin{adjustbox}{width=0.65\textwidth}
    \begin{tabular}{l|ccc|c}
    \toprule
    \textbf{Method}	& AIME24 & AIME25 & MATH500 & Average \\
    \hline      
{\underline{Qwen2.5-Coder-7B-Instruct}} & 7.1 	& 2.5 	&  58.6 &	22.7 	\\
    + ReTool-SFT  &  15.8	& 12.1 	& \textbf{73.9} &	33.9 	\\
    + ReTool-SFT w/ \textsc{CoBe}  &  \textbf{23.3}	&  \textbf{13.8}	& 73.2 & \textbf{36.8}	 	\\
    \hdashline
    + ReTool-RL  &  40.0	& 29.6 	& 84.4  & 51.6	 	 \\
    + ReTool-RL w/ \textsc{CoBe}  &  \textbf{42.1}	& \textbf{31.7} 	& \textbf{84.5}  & \textbf{52.8}	 	\\    
    \hline    
{\underline{Qwen2.5-7B-Instruct}} &  12.1	& 3.3	& 69.0 &	 28.2	\\
    + ReTool-SFT  &  13.3	& 10.4 	& 75.7 &	33.1 	\\
    + ReTool-SFT w/ \textsc{CoBe}  &  \textbf{17.1}	& \textbf{14.2} 	& \textbf{74.5} & \textbf{35.2}	 	\\
    \hdashline
    + ReTool-RL &  32.5 &  24.6 & 77.1 &	44.7 	 \\
    + ReTool-RL w/ \textsc{CoBe}  &  \textbf{40.0}	& \textbf{26.7} 	& \textbf{81.9} &	\textbf{49.5} 	\\
    \hline
{\underline{Qwen3-8B}} & \textbf{43.8} 	&  \textbf{37.9}	& 73.7 &	51.8 	\\
    + ReTool-SFT &  40.8 &  27.3	& \textbf{88.2} &	\textbf{52.1} 	\\
    + ReTool-SFT w/ \textsc{CoBe} &  32.5	& 21.3 	& 81.5 &	45.1 	\\
    \hdashline
    + ReTool-RL  &  \textbf{67.5}	&  \textbf{59.2}	& \textbf{95.4} &	 \textbf{74.0}	 \\
    + ReTool-RL w/ \textsc{CoBe} & 58.8 	& 46.3 	& 91.9 &	65.6 	\\    
    \bottomrule
    \end{tabular}
    \end{adjustbox}
    \end{center}
    \label{tab:main-result}
\end{table}

\paragraph{Result and Discussion.}
As shown in \Cref{tab:main-result}, ReTool improves mathematical reasoning performance on both Qwen2.5-7B-Instruct and Qwen2.5-Coder-7B-Instruct across the SFT and RL stages. Building on this foundation, incorporating \textsc{CoBe} yields additional gains in two of the three models, suggesting that \textsc{CoBe} can effectively complement the ReTool training framework. In contrast, Qwen3-8B does not benefit from \textsc{CoBe} and instead exhibits performance degradation relative to the corresponding ReTool baseline. This discrepancy reveals that the effectiveness of \textsc{CoBe} may depend on model-specific characteristics. We investigate this phenomenon further in \Cref{sec:analysis:qwen3-degrade}.
Examining the training stages separately, augmenting ReTool with \textsc{CoBe} during SFT improves the average performance of Qwen2.5-Coder-7B-Instruct from 33.9 to 36.8 and Qwen2.5-7B-Instruct from 33.1 to 35.2. The gains become more pronounced during RL, where ReTool-RL with \textsc{CoBe} further increases the average score from 51.6 to 52.8 on Qwen2.5-Coder-7B-Instruct and from 44.7 to 49.5 on Qwen2.5-7B-Instruct.
These results suggest that cognitive behaviors can be effectively leveraged at training time through \textsc{CoBe} augmentation, providing additional supervision that improves downstream mathematical reasoning, particularly for Qwen2.5-based models. Further analysis of the observed improvements beyond aggregate performance is provided in \Cref{sec:analysis-beyond-accuracy}.

\section{Analysis}

\subsection{Investigating the Lack of Performance Improvements in Qwen3-8B}
\label{sec:analysis:qwen3-degrade}

\begin{figure}[h!]
    \centering
    \includegraphics[width=0.8\textwidth]{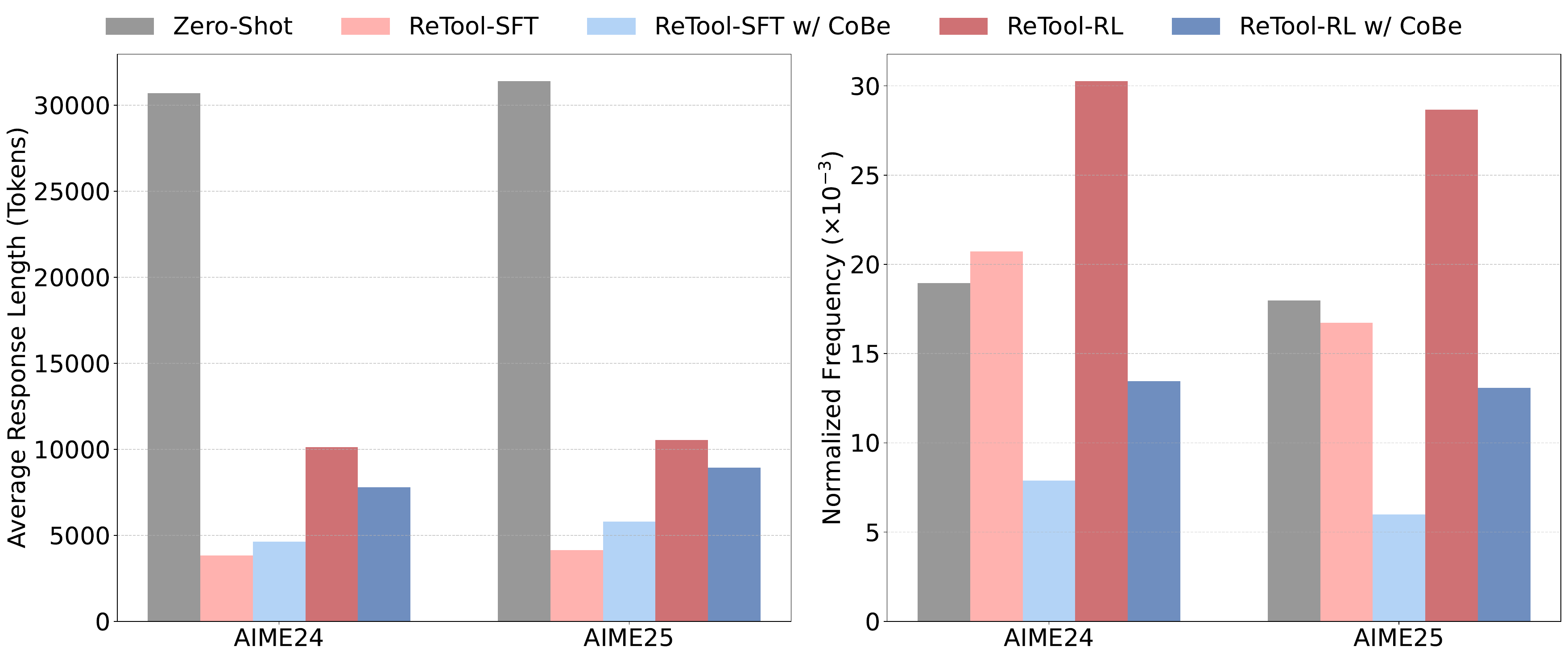}
    \caption{
         Comparative analysis of Qwen3-8B outputs across mathematical benchmarks. (Left) Comparison of average response length between the zero-shot, standard ReTool, and ReTool integrated with \textsc{CoBe}. (Right) Length-normalized frequency of total \emph{crucial tokens}, referenced from Section~\ref{subsec:extrinsic}.
     }
    \label{fig:analysis-qwen3}
\end{figure}

Although Qwen3-8B was explicitly primed with the desired intrinsic property, we verify that \textsc{CoBe}-trained models consistently exhibit the intended behaviors during downstream evaluation (see \Cref{appendix:additional-analysis}). We observe no downstream performance improvement.
We hypothesize that this phenomenon may be confounded by two factors: (1) discrepancies in response-length distributions introduced during SFT, and (2) the insufficient preservation of the extrinsic property.
Regarding response length, the left panel of \Cref{fig:analysis-qwen3} shows that outputs produced by both ReTool-SFT and ReTool-SFT with \textsc{CoBe} are substantially shorter than the corresponding zero-shot generations. While response length itself may not directly determine downstream accuracy, such a substantial distributional shift may negatively affect model behavior. This observation is consistent with the degraded AIME performance of both SFT variants reported in \Cref{tab:main-result}.

We further assess the relationship between performance and the extrinsic property through an analysis of \emph{crucial token} frequency. As illustrated in the right panel of \Cref{fig:analysis-qwen3}, ReTool-RL exhibits the higher length-normalized frequency of \emph{crucial tokens} on both AIME24 and AIME25, aligning with its superior performance. By comparison, ReTool-RL with \textsc{CoBe} yields lower \emph{crucial token} frequencies and correspondingly weaker results.
Overall, this finding highlights that maintaining both intrinsic and extrinsic properties is also important for preserving strong reasoning performance.

To further disentangle these factors, we conduct an additional empirical study using Qwen3-8B-Base instead of the post-trained Qwen3-8B model. We apply both ReTool-SFT and ReTool-SFT with \textsc{CoBe} to the base model in order to isolate the effects.
As demonstrated in \Cref{appendix:additional-results}, incorporating \textsc{CoBe} into the base model setting improves the ReTool-SFT baseline by approximately 7\% on average. In particular, performance increases by 15\% on AIME24 and 35.1\% on AIME25, while performance on MATH500 decreases by only 1.1\%.
These results suggest that \textsc{CoBe} can be beneficial when applied to the base model, but its effectiveness may vary for models that have already undergone extensive post-training. Further investigation is therefore required to understand how cognitive-behavior augmentation interacts with existing model capabilities.

\subsection{The Impact of Cognitive Behaviors on Reasoning With CI Beyond Task Performance} 
\label{sec:analysis-beyond-accuracy}

\begin{figure*}[htbp]
    \centering
    \begin{subfigure}[t]{0.32\textwidth}
        \centering
        \includegraphics[width=\linewidth]{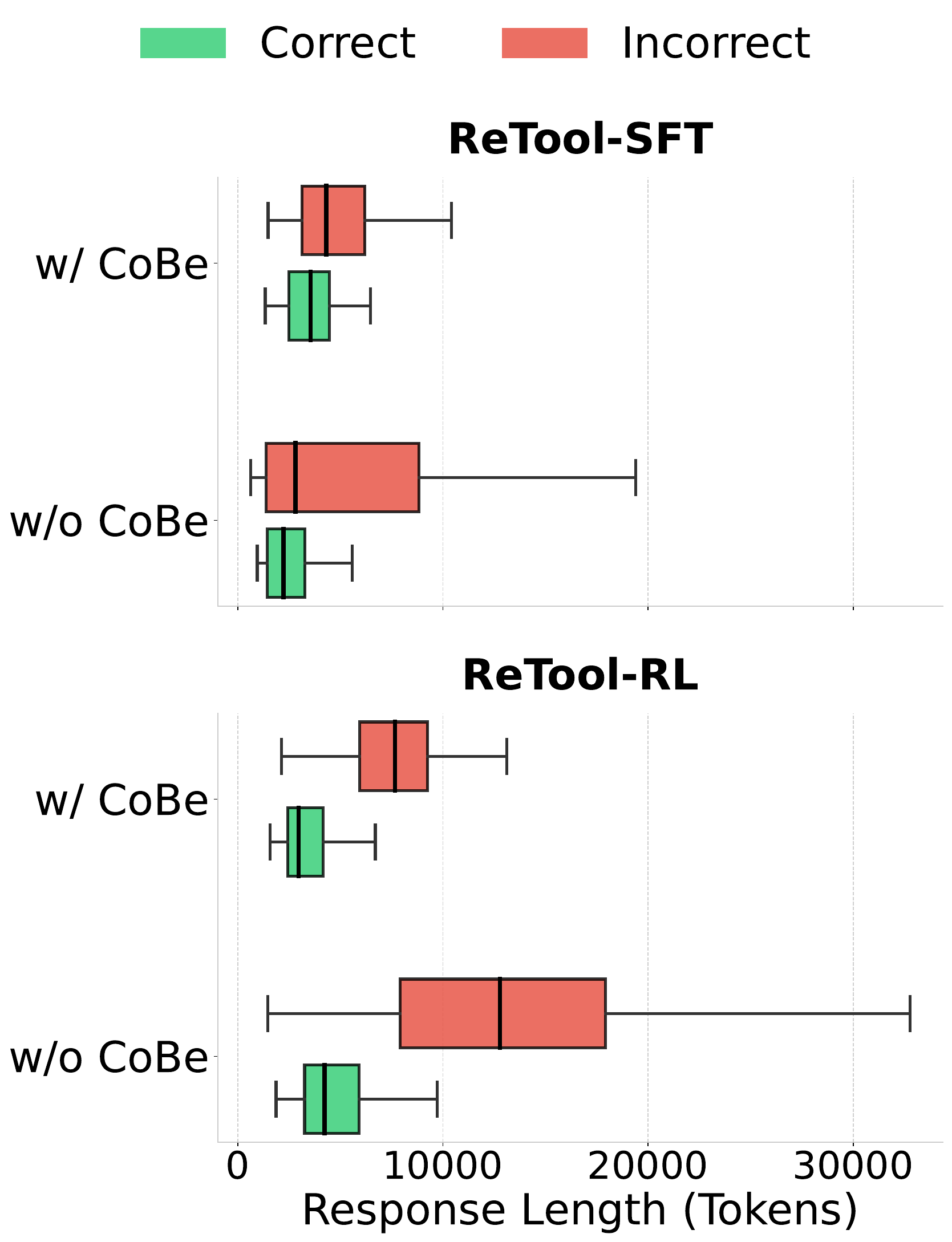}
        \caption{Response lengths.}
        \label{fig:overthinking}
    \end{subfigure}
    \hfill
    \begin{subfigure}[t]{0.64\textwidth}
        \centering
        \includegraphics[width=0.97\linewidth]{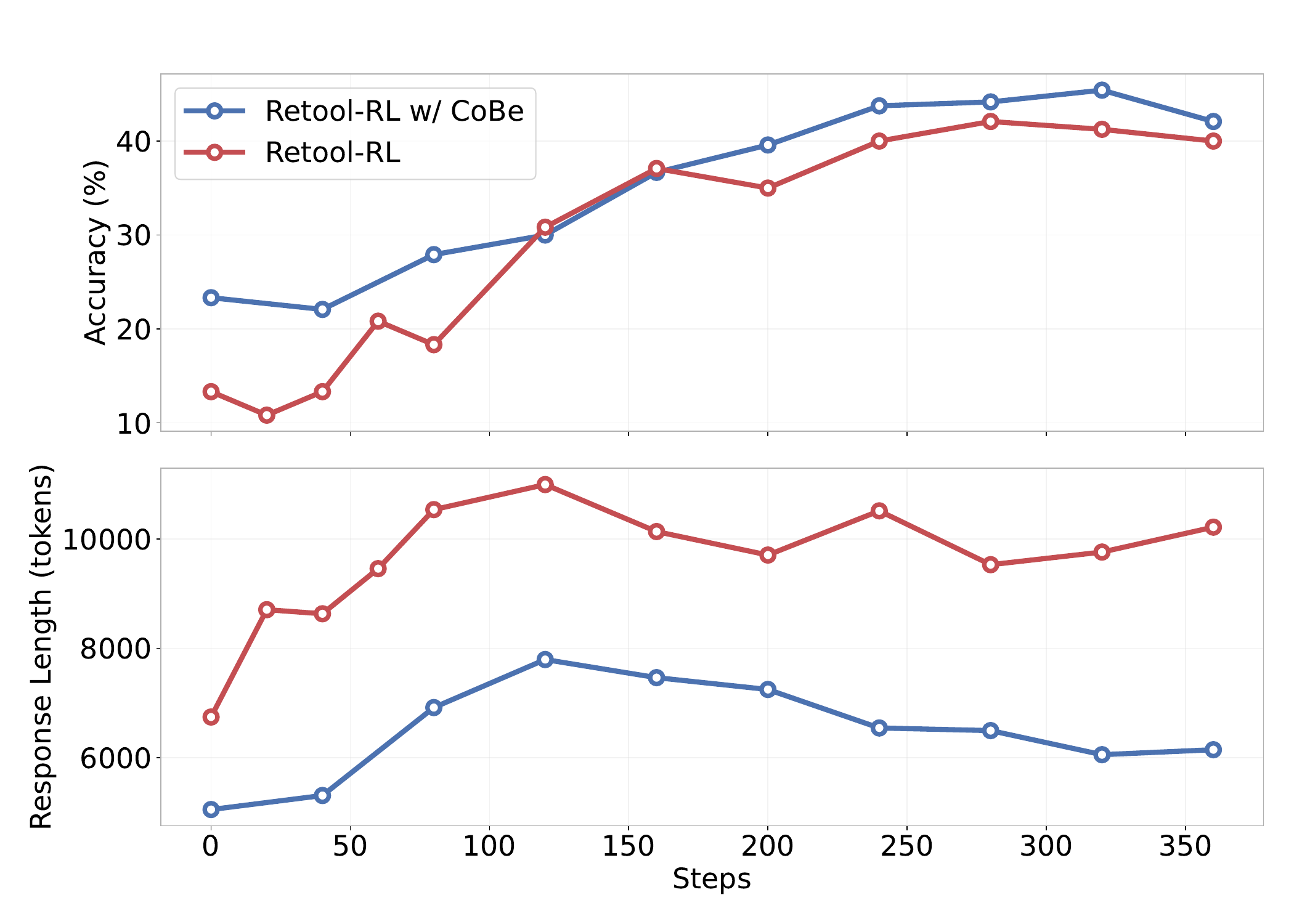}
        \caption{Evolution of accuracy and average response lengths.}
        \label{fig:evolution-accuracy-length}
    \end{subfigure}

    \caption{
   Response length distributions and training dynamics of Qwen2.5-7B-Coder-Instruct on AIME24. \textsc{CoBe} reduces overthinking on incorrect predictions, enhancing token efficiency and accuracy.
    }
    \label{fig:analysis-beyond}
\end{figure*}

Prior work \citep{underthinking} has observed that incorrect reasoning trajectories are often associated with excessively long generations, suggesting that LMs may overthink when faced with difficult problems. Motivated by this insight, we investigate whether incorporating \textsc{CoBe} into the training pipeline can encourage more structured reasoning and thereby mitigate inefficient generation behaviors.
\Cref{fig:analysis-beyond} shows that integrating \textsc{CoBe} consistently reduces response lengths for incorrect samples across both the SFT and RL stages. This trend indicates that \textsc{CoBe} enables the model to reason more efficiently, particularly in cases where the model ultimately fails to reach the correct solution.

We further compare the training dynamics of ReTool-RL with and without \textsc{CoBe} under identical training settings. The \textsc{CoBe}-enhanced variant outperforms the standard ReTool-RL baseline in 8 out of 10 checkpoints on AIME24, while consistently generating shorter responses.
In contrast, standard ReTool-RL exhibits a bias towards longer responses with limited accuracy improvements, reflecting the length bias identified by \citet{drgrpo}.
These results suggest that extending length alone is insufficient for reasoning with a code interpreter; instead, the quality and structure of intermediate reasoning behaviors are also critical.

\section{Conclusion}
Although reasoning with code interpreter is increasingly adopted, the underlying extrinsic \emph{crucial tokens} and intrinsic cognitive behaviors that facilitate effective reasoning remain largely underexplored. To address this, we conduct a systematic study of how these properties manifest across multiple LLMs, focusing on \emph{crucial tokens} and cognitive behaviors such as verification, backtracking, subgoal setting, and backward chaining.
Our results show that both properties can be leveraged to improve downstream performance in certain conditions. In particular, \emph{crucial token} intervention, especially with \emph{let}, enables effective test-time scaling, while incorporating code-specific cognitive behaviors into training improves performance during both supervised fine-tuning and reinforcement learning in most cases. Beyond accuracy gains, these behaviors also improve token efficiency and reduce overthinking in incorrect responses, highlighting the importance of intrinsic behavior rather than simply extending generation length. At the same time, the performance degradation observed in a certain specialized model suggests that the effectiveness of behavioral augmentation depends on its interaction with existing model capabilities, motivating further investigation into factors that constrain gains in advanced reasoning models.

\subsubsection*{Limitations and Potential Risks}

Although this work provides a systematic characterization of extrinsic and intrinsic properties in code-interpreter-based reasoning, several limitations should be acknowledged. First, both the evaluation process and data synthesis pipeline rely on high-capacity Oracle LLMs, which may introduce model-specific biases and variability in generated outputs. Moreover, our interventions investigate these properties independently: training primarily targets intrinsic behaviors (\textsc{CoBe}), whereas test-time scaling focuses on \emph{crucial tokens} associated with extrinsic properties. As demonstrated in our analysis of Qwen3-8B, reductions in \emph{crucial token} frequency can adversely affect task performance even when intrinsic reasoning behaviors are maintained. Consequently, this study does not provide a unified framework that jointly leverages both intrinsic and extrinsic properties.
Second, the observed findings may not readily generalize to broader tool-augmented agent settings, such as web browsing, multi-tool orchestration, or open-ended API interactions. In addition, our conclusions are derived from a specific set of publicly available datasets and evaluation frameworks, which may limit their applicability to other domains and tasks.
Finally, this study involves the execution of LLM-generated code as part of the evaluation process. While all executions are performed within controlled sandbox environments and are unlikely to produce harmful effects, the possibility of unintended or unanticipated behaviors arising from model-generated code cannot be entirely eliminated.



\subsubsection*{Acknowledgments}
This research is supported by the National Research Foundation, Singapore, under its National Large Language Models Funding Initiative. Any opinions, findings and conclusions or recommendations expressed in this material are those of the author(s) and do not reflect the views of National Research Foundation, Singapore.
In addition, the computing resources used in this project were supported by the WangchanX program, a research and development initiative by PyThaiNLP and VISTEC-DEPA Thailand AI Research Institute.
This work was further supported by the ThaiLLM collaboration, funded by the Digital Economy and Society (DE) Development Fund of the Ministry of Digital Economy and Society, Thailand.

\bibliography{tmlr}
\bibliographystyle{tmlr}

\appendix

\section{Sample of SFT dataset}
\label{appendx:example}

Figure~\ref{fig:reason-with-ci} illustrates the surface example of the SFT dataset from ReTool \cite{retool}, which consists of user prompts, assistant responses with Python scripts, and code interpreter results.

\begin{figure}[htbp]
\centering
\begin{tcolorbox}[
    enhanced,
    colback=white,
    colframe=black,
    arc=0mm,
    boxrule=0.7pt,
    left=3mm, right=3mm, top=2mm, bottom=2mm,
]
    \small
    \textbf{User} \\
    There are 152 students at Dala High School. Assume the following: \\
    - 100 students take a Math class \\
    - 94 take a Science class \\
    - 57 take an English class. \\
    - 73 take Math and Science \\
    - 24 take Math and English \\
    - 27 take Science and English. \\
    - 22 take Math, Science, and English. \\
    How many students take neither a Math class nor a Science class nor an English class? \\
    Remember to place the final answer in the format: 
    \texttt{\textbackslash boxed\{`The final answer goes here.'\}}

    \vspace{1.5mm} \hrule \vspace{1.5mm}
    \textbf{Assistant} \\
    Okay, I need to find the number of students taking none of the classes. The total is 152. I will use the principle of inclusion-exclusion (PIE) for three sets: $|M \cup S \cup E| = |M| + |S| + |E| - (|M \cap S| + |M \cap E| + |S \cap E|) + |M \cap S \cap E|$. Then, subtract that from the overall total.
    
    \vspace{1mm}
    \begin{tcolorbox}[colback=gray!5, colframe=gray!20, boxrule=0.5pt, arc=1mm, left=2pt, top=2pt, bottom=2pt]
    \begin{lstlisting}[language=Python, basicstyle=\footnotesize\ttfamily]
M, S, E = 100, 94, 57
M_S, M_E, S_E = 73, 24, 27
M_S_E = 22
total_in_any = M + S + E - M_S - M_E - S_E + M_S_E
print(152 - total_in_any)
    \end{lstlisting}
    \end{tcolorbox}
    \texttt{<interpreter>3</interpreter>}

    \vspace{1.5mm}
    The code returns 3. To verify, I will break down each disjoint category (e.g., "Only Math") to ensure the total population sum is consistent.
    
    \vspace{1mm}
    \begin{tcolorbox}[colback=gray!5, colframe=gray!20, boxrule=0.5pt, arc=1mm, left=2pt, top=2pt, bottom=2pt]
    \begin{lstlisting}[language=Python, basicstyle=\footnotesize\ttfamily]
only_M_S, only_M_E, only_S_E = 73-22, 24-22, 27-22
only_M = 100 - only_M_S - only_M_E - 22
only_S = 94 - only_M_S - only_S_E - 22
only_E = 57 - only_M_E - only_S_E - 22
total_verify = only_M + only_S + only_E + only_M_S + only_M_E + only_S_E + 22
print(total_verify)
    \end{lstlisting}
    \end{tcolorbox}
    \texttt{<interpreter>149</interpreter>}

    \vspace{1.5mm}
    The verification total is 149, so students not taking any classes are 152 - 149 = 3, confirming the initial result. \texttt{\textbackslash boxed\{3\}}

\end{tcolorbox}
\caption{Illustrative multi-turn reasoning trace featuring interleaved code execution. This sample is sourced from the ReTool cold-start supervised fine-tuning data \citep{retool}.}
\label{fig:reason-with-ci}
\end{figure}


\section{Probing Code-Specific Cognitive Behaviors in Reasoning with CI}
\label{appendx:probing-code}

In preliminary experiments, we found that the cognitive-behavior prompts proposed by \citet{cognitive-behavior} do not readily transfer to the CI-based reasoning setting. For instance, as illustrated in \Cref{fig:example-prelim-cognitive-fail}, the original framework fails to identify any of the four cognitive behaviors in a representative CI-based reasoning trajectory.
To better capture code-specific cognitive behaviors, we adapt the prompting framework for the CI-based reasoning regime. Consistent with prior work, each behavior is evaluated independently. Under the revised framework, the same trajectory is identified as exhibiting both subgoal setting and backtracking.
The prompts used for backtracking and backward chaining are presented in \Cref{fig:probe-cobe-prompt-backtrack-and-backward}, while those for subgoal setting and verification are provided in \Cref{fig:probe-cobe-prompt-subgoal-and-verification}.

\begin{figure}[htbp]
\centering
\begin{tcolorbox}[
    enhanced,
    colback=white,
    colframe=black,
    arc=0mm,
    boxrule=0.7pt,
    left=3mm, right=3mm, top=2mm, bottom=2mm,
]
    \scriptsize
    \textbf{User} \\
    Xiao Wang is a librarian responsible for shelving 6 books labeled A, B, C, D, E, F on a shelf with positions 1 to 6. \\ \\The constraints are: \\1. Book D must be to the left of book F. \\2. Book D must be adjacent to book A. \\3. Book F must be to the left of book C. \\4. Book C cannot be placed in position 1. \\5. Book B cannot be placed in position 1. \\6. Book C must be to the right of book B. \\7. Book A must be to the left of book C. \\ \\Question: \\Based on the above conditions, please arrange the 6 books from left to right on the bookshelf, i.e. positions 1 through 6. Return your answer with the format <<< a list of strings >>>, one solution is enough.for example: \\<<<["A", "B", "C"]>>> \\Your answer:
    \texttt{\textbackslash boxed\{`The final answer goes here.'\}}

    \vspace{1.5mm} \hrule \vspace{1.5mm}
    \textbf{Assistant} \\
    To solve this problem, we need to find a valid arrangement of the books A, B, C, D, E, and F that satisfies all the given constraints. Let's break down the constraints step by step: \\ \\1. Book D must be to the left of book F. \\2. Book D must be adjacent to book A. \\3. Book F must be to the left of book C. \\4. Book C cannot be placed in position 1. \\5. Book B cannot be placed in position 1. \\6. Book C must be to the right of book B. \\7. Book A must be to the left of book C. \\ \\Given these constraints, we can use a brute-force approach to try all possible permutations of the books and check which one satisfies all the conditions. Let's implement this in Python.
    
    \vspace{1mm}
    \begin{tcolorbox}[colback=gray!5, colframe=gray!20, boxrule=0.5pt, arc=1mm, left=2pt, top=2pt, bottom=2pt]
    \begin{lstlisting}[language=Python, basicstyle=\scriptsize\ttfamily]
from itertools import permutations

# Define the books
books = ['A', 'B', 'C', 'D', 'E', 'F']

# Function to check if a permutation satisfies all constraints
def is_valid(arrangement):
    # Constraint 1: Book D must be to the left of book F
    if arrangement.index('D') > arrangement.index('F'):
        return False
    # Constraint 2: Book D must be adjacent to book A
    if abs(arrangement.index('D') - arrangement.index('A')) != 1:
        return False
    # Constraint 3: Book F must be to the left of book C
    if arrangement.index('F') > arrangement.index('C'):
        return False
    # Constraint 4: Book C cannot be placed in position 1
    if arrangement[0] == 'C':
        return False
    # Constraint 5: Book B cannot be placed in position 1
    if arrangement[0] == 'B':
        return False
    # Constraint 6: Book C must be to the right of book B
    if arrangement.index('C') < arrangement.index('B'):
        return False
    # Constraint 7: Book A must be to the left of book C
    if arrangement.index('A') > arrangement.index('C'):
        return False
    return True

# Generate all permutations and find a valid one
for arrangement in permutations(books):
    if is_valid(arrangement):
        print("<<<{}>>>".format(arrangement))
        break
    \end{lstlisting}
    \end{tcolorbox}

\end{tcolorbox}
\caption{Partial response generated by Qwen2.5-7B-Instruct on \texttt{permutations-and-combinations} task.}
\label{fig:example-prelim-cognitive-fail}
\end{figure}




\begin{figure}[htbp]
\centering
\begin{tcolorbox}[
    enhanced,
    colback=white,
    colframe=black,
    arc=0mm,
    boxrule=0.7pt,
    left=3mm, right=3mm, top=2mm, bottom=2mm,
]
    \scriptsize
    \textbf{Backtracking}
    \vspace{1.5mm} \hrule \vspace{1.5mm}
\# Task Description\\
You will be provided with code reasoning or explanation text. Your task is to detect whether the text includes algorithmic backtracking behavior — where the writer explicitly describes attempting one path or strategy in code, realizing it fails, and then reversing or undoing that choice to try another.\\
\\
Focus on algorithmic backtracking, not surface-level rewordings or corrections.\\
\\
Backtracking behavior may include:\\
- Describing recursive or trial-based algorithms that explore options and abandon invalid ones\\
- Reverting a previous decision in code (e.g., “remove”, “undo”, or “pop” a value)\\
- Trying all options in a loop or recursion, and skipping ones that violate constraints\\
- Returning early or continuing to next candidate when current state is invalid\\
- Mentioning backtracking explicitly as part of the approach\\
\\
If you find any backtracking steps, count them and provide the count between the tags `<count>` `</count>`. If there is no backtracking behavior, provide a count of 0 as `<count>0</count>`.\\
\\
\# Task Format\\
Format your response in markdown as follows:\\
\\
\#\# Thoughts  \\
{[Brief description of what algorithmic backtracking behavior was noticed and where]}\\
\\
\#\# Does backtrack?  \\
{[yes/no]}\\
\\
\#\# Number of backtrack steps  \\
<count> [number] </count>\\
\\
\# Task to evaluate for backtracking  \\
\{response\}\\
\\
\# Response

\end{tcolorbox}
\begin{tcolorbox}[
    enhanced,
    colback=white,
    colframe=black,
    arc=0mm,
    boxrule=0.7pt,
    left=3mm, right=3mm, top=2mm, bottom=2mm,
]
    \scriptsize
    \textbf{Backward Chaining}
    \vspace{1.5mm} \hrule \vspace{1.5mm}
\# Task Description\\
You will be provided with code reasoning or explanation text. Your task is to detect whether the text includes backward chaining behavior — where the writer starts from a desired goal, final output, or condition, and reasons *backward* to determine what inputs, conditions, or steps would achieve it.\\
\\
This behavior is common in constraint-solving, logic programming, test-driven design, and debugging.\\
\\
Backward chaining in code reasoning may include:\\
- Starting from the required output and asking what input would produce it\\
- Working backward from test results, outputs, or constraints to figure out the logic\\
- Describing a condition (e.g. “to return True…”) and identifying how to meet it\\
- Setting up reverse planning steps (e.g. “to reach this final state, we need…”)\\
- Using final state or assertions as anchors for inferring code paths\\
\\
If you find any such backward-chaining steps, count them and provide the count between the tags `<count>` `</count>`. If there are no backward-chaining behaviors, provide a count of 0 as `<count>0</count>`.\\
\\
\# Task Format\\
Format your response in markdown as follows:\\
\\
\#\# Thoughts  \\
{[Brief description of what backward chaining behavior was noticed and where]}\\
\\
\#\# Does backward chaining?  \\
{[yes/no]}\\
\\
\#\# Number of backward chaining instances  \\
<count> [1/2/...] </count>\\
\\
\# Task to evaluate for backward chaining  \\
\{response\}\\
\\
\# Response

\end{tcolorbox}
\caption{Prompt for analyzing backtracking and backward chaining.}
\label{fig:probe-cobe-prompt-backtrack-and-backward}
\end{figure}

\begin{figure}[htbp]
\centering
\begin{tcolorbox}[
    enhanced,
    colback=white,
    colframe=black,
    arc=0mm,
    boxrule=0.7pt,
    left=3mm, right=3mm, top=2mm, bottom=2mm,
]
    \scriptsize
    \textbf{Subgoal Setting}
    \vspace{1.5mm} \hrule \vspace{1.5mm}
\# Task Description\\
You will be provided with an explanation or reasoning related to code problem-solving. Your task is to detect whether the writer engages in explicit subgoal setting — where they break down a complex coding task into smaller, intermediate objectives.\\
\\
Subgoal setting in code reasoning may include:\\
- Listing or sequencing intermediate steps to reach the final solution\\
- Splitting a large function or algorithm into subtasks or helper routines\\
- Phrases indicating ordered steps (e.g., “first we…”, “then we…”, “finally…”)\\
- Planning ahead by structuring work into stages such as parsing input, preprocessing data, then computing output\\
- Describing recursive decomposition, e.g., “we solve the base case, then handle the recursive step”\\
\\
If you find any subgoal-setting steps, count them and provide the count between the tags `<count>` `</count>`. If the text does not contain any subgoal setting, provide a count of 0 as `<count>0</count>`.\\
\\
\# Task Format\\
Format your response in markdown as follows:\\
\\
\#\# Thoughts  \\
{[Brief description of what subgoal-setting behavior was noticed and where]}\\
\\
\#\# Does subgoal setting?  \\
{[yes/no]}\\
\\
\#\# Number of subgoal setting steps  \\
<count> [number] </count>\\
\\
\# Task to evaluate for Subgoal Setting  \\
\{response\}\\
\\
\# Response\\

\end{tcolorbox}
\begin{tcolorbox}[
    enhanced,
    colback=white,
    colframe=black,
    arc=0mm,
    boxrule=0.7pt,
    left=3mm, right=3mm, top=2mm, bottom=2mm,
]
    \scriptsize
    \textbf{Verification}
    \vspace{1.5mm} \hrule \vspace{1.5mm}

\# Task Description\\
You will be provided with a piece of code and an explanation or reasoning about how that code solves a problem. Your task is to identify whether the explanation contains any verification steps that check or validate the code’s correctness. Since the code cannot actually be run here, look for logical checks, sample input walkthroughs, and sanity checks in the reasoning.\\
\\
Verification steps may include:\\
- Running the code mentally with sample inputs to see if outputs match expectations.\\
- Comparing the code’s output or behavior to the problem requirements or expected results.\\
- Checking edge cases (for example, empty input, very large values, negative values) in the reasoning.\\
- Inspecting function signatures, parameter types, or return types to ensure they match the specifications.\\
- Confirming that loops terminate or that recursion depth stays within safe limits given the constraints.\\
- Analyzing time or space complexity to ensure it is acceptable for the input size.\\
- Ensuring that all conditional branches and special cases in the code are considered.\\
- Verifying loop invariants or assumptions made in the code logic.\\
\\
If you find any verification steps, count them and provide the count between the tags <count> </count>. If the text does not contain any verification steps, provide a count of 0 as <count>0</count>.\\
\\
\# Task Format\\
Format your response in markdown as follows:\\
\\
\#\# Thoughts\\
{[Brief description describing what reasoning-based verification behaviors were detected.]}\\
\\
\#\# Does verification?\\
{[yes/no]}\\
\\
\#\# Number of verification steps\\
<count> [number] </count>\\
\\
\#\# Task to evaluate for Verification\\
\{response\}\\
\\
\#\# Response

\end{tcolorbox}
\caption{Prompt for analyzing subgoal setting and verification.}
\label{fig:probe-cobe-prompt-subgoal-and-verification}
\end{figure}

\section{Synthesize Code-Specific Cognitive Behaviors}
\label{appendix:syn-cobe}

\begin{figure*}[tb]
    \centering
    \begin{subfigure}[t]{0.48\textwidth}
        \centering
        \includegraphics[width=\linewidth]{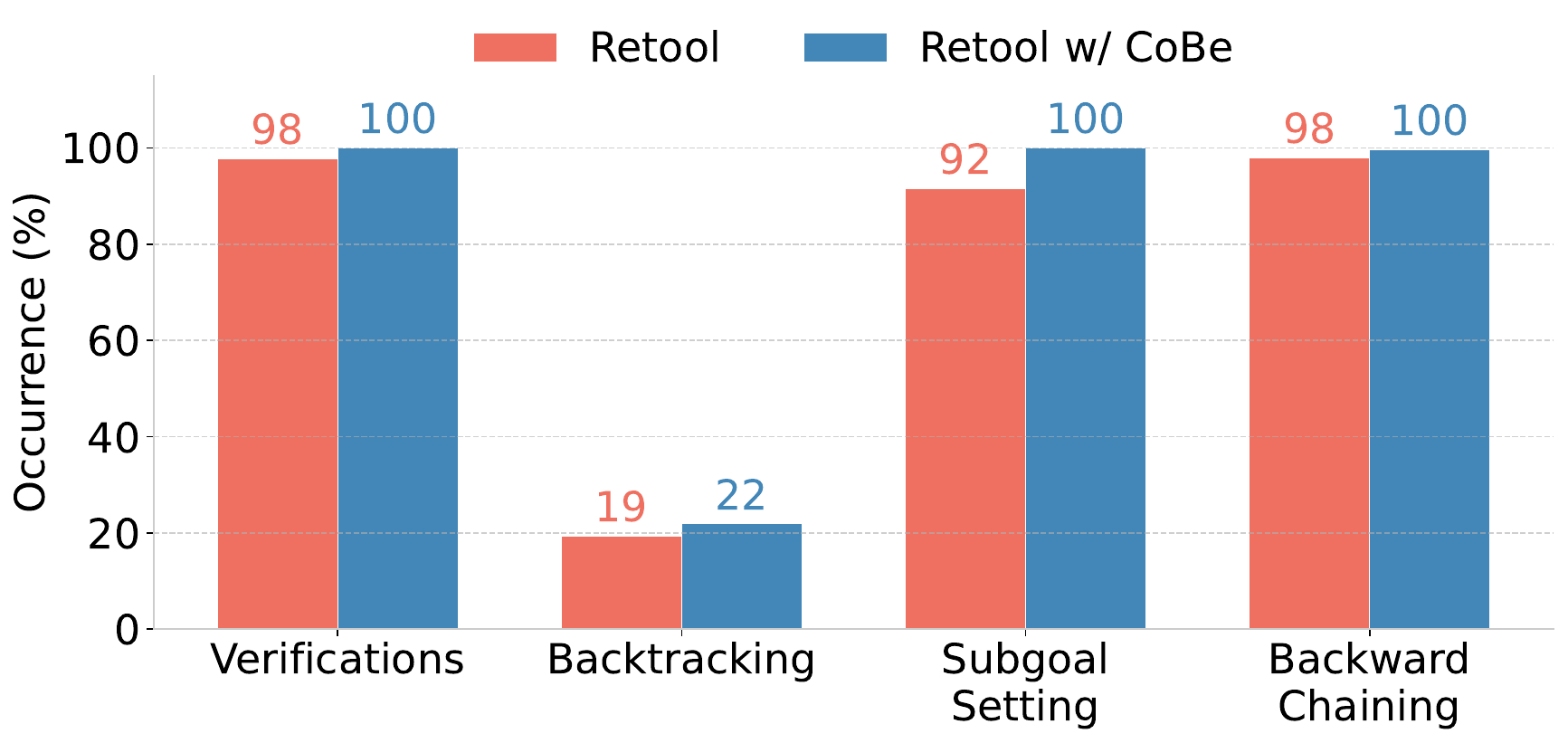}
        \caption{Code-specific cognitive behaviors.}
        \label{fig:stat-retool-sft-with-cobe}
    \end{subfigure}
    \hfill
    \begin{subfigure}[t]{0.48\textwidth}
        \centering
        \includegraphics[width=\linewidth]{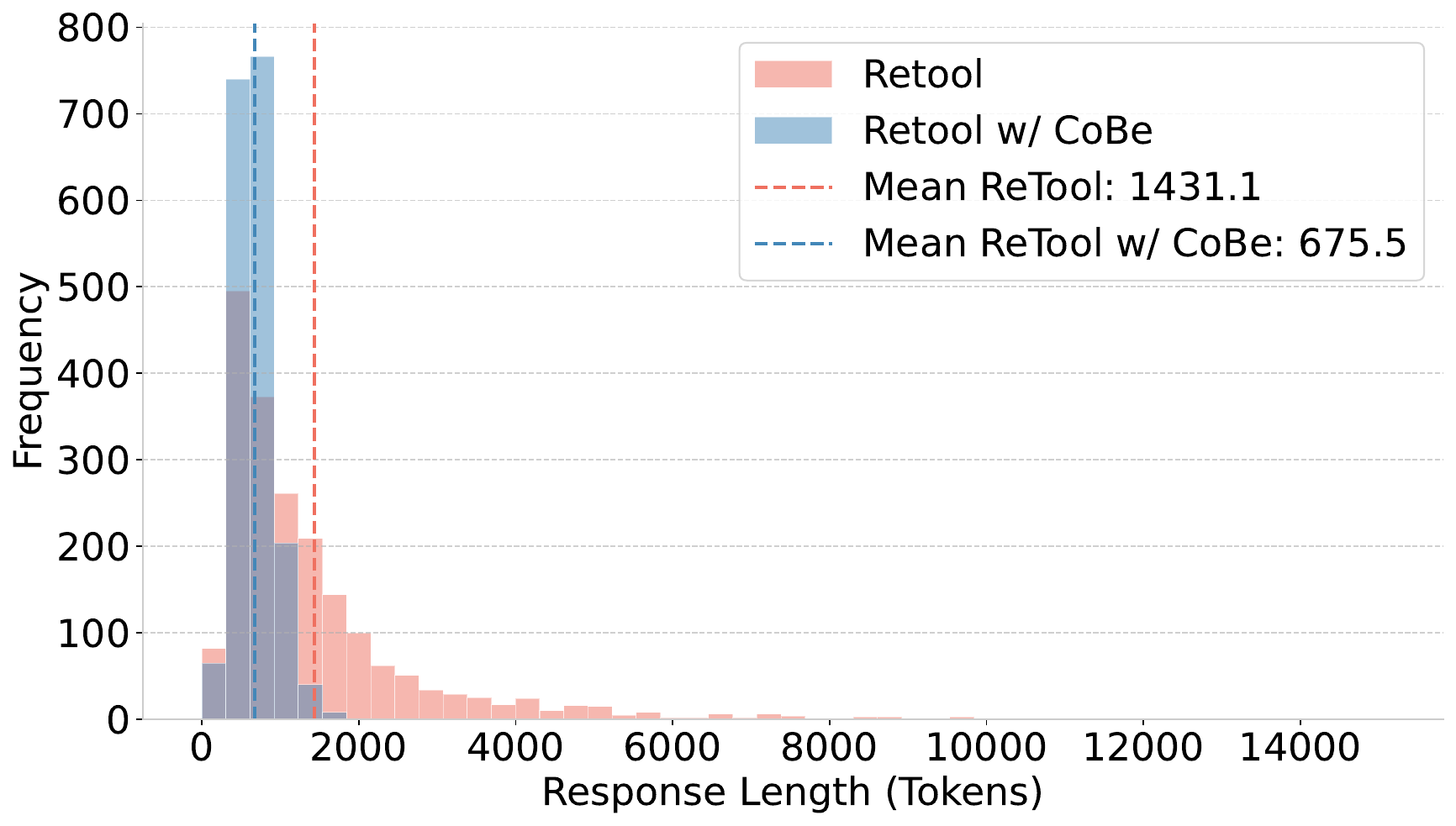}
        \caption{Response length comparisons.}
        \label{fig:training-data-length}
    \end{subfigure}

    \caption{
    Quantitative comparison between standard ReTool SFT data and ReTool augmented with CoBe across two characteristics: the prevalence of cognitive behaviors and response length distributions.
    }
    \label{fig:cobe-characters}
\end{figure*}

To augment ReTool with \textsc{CoBe}, we follow the data synthesis procedure of \citet{cognitive-behavior}. Specifically, we construct a new dataset, $\mathcal{D}_{\text{CoBe}}=(\mathbf{x}, \mathbf{y}_{\text{CoBe}})$, from the original ReTool SFT dataset\footnote{\url{https://huggingface.co/datasets/JoeYing/ReTool-SFT}} $\mathcal{D}=(\mathbf{x}, \mathbf{y})$ by prompting Claude Sonnet 4.5 with both the original trajectory $y$ and a cognitive-behavior prompt $p_{\text{CoBe}}$ (\Cref{fig:syn-cobe-prompt}). The synthesized trajectory is generated as
$
\mathbf{y}_{\text{CoBe}} \sim \text{LLM}(\mathbf{x}, \mathbf{y}; p_{\text{CoBe}}, \text{CI}).
$
To ensure correctness, we retain only synthesized trajectories whose final answers match the corresponding gold labels, yielding 1,915 samples from the original 2,000 instances.
We further compare the prevalence of cognitive behaviors between the original ReTool-SFT trajectories and the \textsc{CoBe}-augmented trajectories in \Cref{fig:stat-retool-sft-with-cobe}. The results indicate substantial increases in the targeted behaviors, with verification, subgoal decomposition, and backward chaining appearing in all synthesized trajectories, while backtracking increases by three percentage points. The distribution of response lengths is shown in \Cref{fig:training-data-length}.


\begin{figure}[htbp]
\centering
\begin{tcolorbox}[
    enhanced,
    colback=white,
    colframe=black,
    arc=0mm,
    boxrule=0.7pt,
    left=3mm, right=3mm, top=2mm, bottom=2mm,
]
    \small
    You are a helpful AI assistant specialized in solving complex mathematical problems with transparent, verifiable computation.\\
    \newline
    You will be given:\\
    1. A complex math problem\\
    2. A gold (reference) solution that includes:\\
    \quad - Natural-language reasoning\\
    \quad - Executable Python code enclosed in \texttt{<code>...</code>}\\
    \quad - Execution outputs wrapped in \texttt{<interpreter>...</interpreter>}\\
    \quad - A final boxed answer in \texttt{<answer> \textbackslash boxed\{\dots\} </answer>} format\\
    \newline
    Task: Generate a new solution trajectory that correctly solves the problem and reaches the same final answer, using structured problem-solving and executable code.\\
    \newline
    You must demonstrate goal-directed problem solving by:\\
    - Decomposing the problem into sub-goals.\\
    - Solving sub-goals to reach the final objective.\\
    - Verifying intermediate results.\\
    - Backtracking and revising earlier steps if needed.\\
    - Working backward from the final goal (of the given problem) if it makes things easier.\\
    \newline
    You should follow the same format as the gold solution, including:\\
    - Natural-language reasoning in \texttt{<think>...</think>} tags\\
    - Executable Python code enclosed in \texttt{<code>...</code>} tags\\
    - Execution outputs from interpreter wrapped in \texttt{<interpreter>...</interpreter>}\\
    - A final answer in \texttt{<answer> \textbackslash boxed\{\dots\} </answer>}\\
    \newline
    \#\#\# Original Conversation\\
    \texttt{<user> \{original\_prompt\} </user>}\\
    \texttt{<bot> \{original\_trajectory\} </bot>}
\end{tcolorbox}
\caption{Prompt for synthesizing a new response with \textsc{CoBe}.}
\label{fig:syn-cobe-prompt}
\end{figure}

\section{Training Configurations and Computational Cost}
\label{appendix:training-config}

All experiments were conducted using the training scripts provided in the official repository\footnote{\url{https://github.com/verl-project/verl-recipe/tree/main/retool}}.
For supervised fine-tuning, models were trained for six epochs with a batch size of 32 and a learning rate of $10^{-6}$. The maximum sequence length was set to 16,384 tokens, and Fully Sharded Data Parallel (FSDP) was employed to enable efficient distributed training.
For reinforcement learning, we adopted the GRPO algorithm with an actor learning rate of $10^{-6}$. The training was performed using a batch size of 64. To support long-chain-of-thought reasoning, the maximum prompt length and response length were set to 2,048 and 16,384 tokens, respectively. During rollout, 16 responses were sampled for each prompt. No KL-divergence regularization term was applied throughout training.
The total computational expenditure, including both training and evaluation, amounted to approximately 1,608 GPU hours on NVIDIA A100 (80GB) GPUs.

\section{Additional Results}
\label{appendix:additional-results}

The results are consistent with our main findings. \textsc{CoBe} improves performance on Qwen2.5-3B-Instruct and Qwen3-8B-Base, suggesting that cognitive-behavior augmentation is generally beneficial. In contrast to Qwen3-8B-Instruct, the gains observed on Qwen3-8B-Base indicate that the degradation is likely related to post-training characteristics rather than the cognitive behaviors themselves.

\begin{table}[htbp]
    \caption{Additional evaluation results of \textsc{CoBe} on a smaller model and a base model.}
    \begin{center}
    \begin{tabular}{l|ccc|c}
    \toprule
    \textbf{Method}	& AIME 24 & AIME25 & MATH500 & Average \\
    \hline
{\underline{Qwen2.5-3B-Instruct}} &  12.1	& 3.3	& 69.0 &	 28.2	\\
    + ReTool-SFT  &  13.3	& 10.4 	& 75.7 &	33.1 	\\
    + ReTool-SFT w/ CoBe  &  \textbf{17.1}	& \textbf{14.2} 	& \textbf{74.5} & \textbf{35.2}	 	\\
    \hline
{\underline{Qwen3-8B-Base}} & - & - & - &	- 	\\
    + ReTool-SFT & 25.4  & 17.1 	& \textbf{85.0} & 42.5 \\
    + ReTool-SFT w/ CoBe & \textbf{29.2} & \textbf{23.0} & 84.1 & \textbf{45.4} \\
    \bottomrule
    \end{tabular}
    \end{center}
    \label{tab:appendix-additional-results}
\end{table}

\section{Additional Analysis}
\label{appendix:additional-analysis}

\begin{figure}[htbp]
    \centering\includegraphics[width=0.8\linewidth]{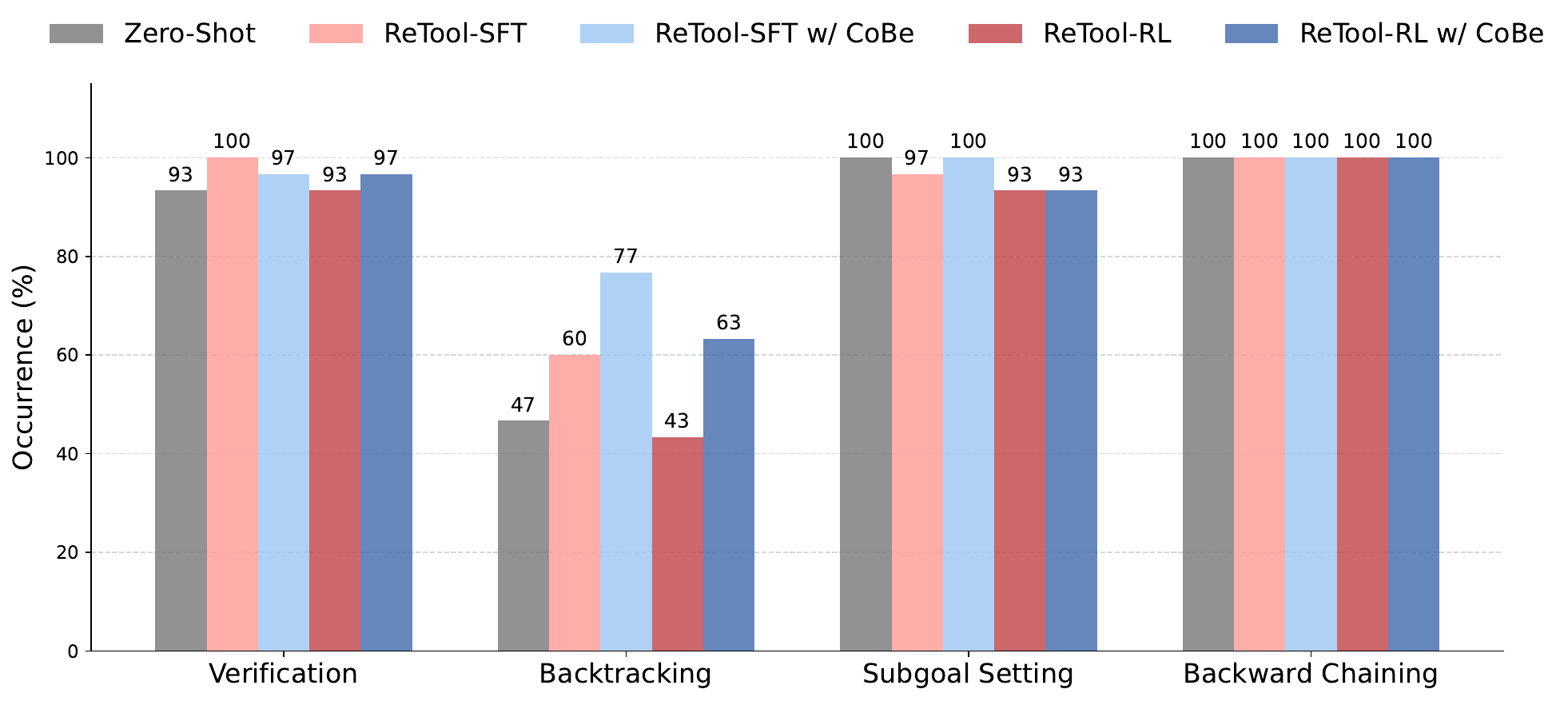}
    \caption{
        Cognitive behaviors of Qwen3-8B under ReTool training with and without \textsc{CoBe} on AIME24.
     }
    \label{fig:beh-aime24}
\end{figure}

\begin{figure}[htbp]
    \centering\includegraphics[width=0.8\linewidth]{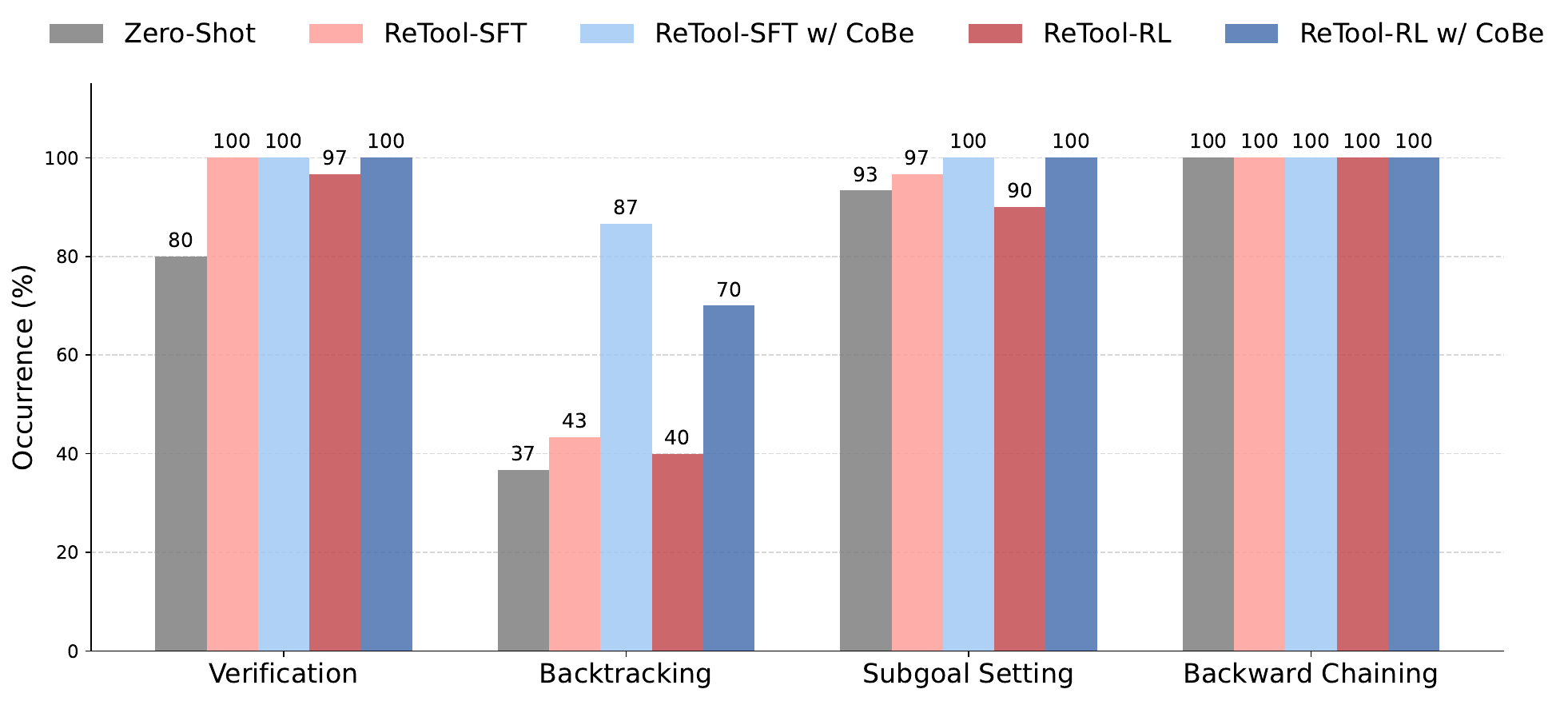}
    \caption{
        Cognitive behaviors of Qwen3-8B under ReTool training with and without \textsc{CoBe} on AIME25.
     }
    \label{fig:beh-aime25}
\end{figure}

Figure~\ref{fig:beh-aime24} and Figure~\ref{fig:beh-aime25} show that ReTool with CoBe consistently induces the desired behaviors during evaluation, except in the SFT variant for verification on AIME 2024, where CoBe is 3\% lower.

\end{document}